\documentclass{article}

\usepackage[letterpaper,top=2cm,bottom=2cm,left=3cm,right=3cm,marginparwidth=1.75cm]{geometry}

\usepackage{amssymb}
\usepackage{siunitx}
\usepackage[linesnumbered,ruled,vlined]{algorithm2e}
\usepackage{amsmath}
\usepackage{graphicx}
\usepackage{newtxtext,newtxmath} % Times New Roman 风格
\usepackage{authblk} % 导入包
\numberwithin{equation}{section}
\usepackage[colorlinks,
linkcolor = blue,	
anchorcolor = blue,
urlcolor = blue,	
citecolor = blue,
]{hyperref}

\usepackage{booktabs}
\usepackage{diagbox}

\title{Data Denoising and Derivative Estimation for Data-Driven Modeling of Nonlinear Dynamical Systems}

\author[1,2,3]{Jiaqi Yao\thanks{Corresponding Author: \texttt{jiaqiyao2030@u.northwestern.edu}}}
\author[1]{Lewis Mitchell}
\author[1]{John Maclean}
\author[2]{Hemanth Saratchandran}

\affil[1]{School of Computer and Mathematical Sciences, The University of Adelaide, 

Adelaide, 5005, South Australia, Australia}

\affil[2]{Australian Institute for Machine Learning, The University of Adelaide, 

Adelaide, 5005, South Australia, Australia}

\affil[3]{Department of Biomedical Engineering, Northwestern University, 

Evanston, 60208, Illinois, USA}

\date{\vspace{-5ex}}

\begin{document}
\maketitle

\begin{abstract}
Data-driven modeling of nonlinear dynamical systems is often hampered by measurement noise. We propose a denoising framework, called Runge-Kutta and Total Variation Based Implicit Neural Representation (RKTV-INR), that represents the state trajectory with an implicit neural representation (INR) fitted directly to noisy observations. Runge–Kutta integration and total variation are imposed as constraints to ensure that the reconstructed state is a trajectory of a dynamical system that remains close to the original data. The trained INR yields a clean, continuous trajectory and provides accurate first-order derivatives via automatic differentiation. These denoised states and derivatives are then supplied to Sparse Identification of Nonlinear Dynamics (SINDy) to recover the governing equations. Experiments demonstrate effective noise suppression, precise derivative estimation, and reliable system identification.
\end{abstract}

\section{Introduction}

Data-driven modeling of nonlinear dynamical systems has become a rapidly expanding research area \cite{ghadami2022data,north2023review,chen2021physics,zhang_subtsbr_2021}. Many real-world phenomena (e.g., the spread of infectious diseases \cite{he2020seir}) are naturally described by dynamical systems. Accurate forecasting of their future behaviour is valuable across domains and becomes feasible when the governing dynamics can be inferred directly from data. A range of techniques pursue this goal, including dynamic mode decomposition (DMD) \cite{tu2013dynamic}, physics-informed neural networks \cite{raissi2019physics}, random feature maps \cite{2023Random}, and graph-Laplacian methods \cite{yamada2024spatial}. In this work, we focus on sparse identification of nonlinear dynamics (SINDy) \cite{brunton2016discovering,viknesh2024adam,fukami2021sparse,carderera2021cindy}.

A well-known limitation across these methods is reduced accuracy in the presence of noise in real-world measurements \cite{garcia2015dealing}. Many algorithms are developed under an assumption of near-perfect data, which is rarely met in practice. This vulnerability has been documented for DMD \cite{wu2021challenges,dawson2016characterizing}, physics-informed neural networks \cite{satyadharma2024assessing,zou2025uncertainty}, and SINDy \cite{wentz2023derivative,rudy2019deep}.

A substantial body of work seeks to improve robustness to noise \cite{hemati2017biasing,thanasutives2023noise}, yet reliably mitigating its effects remains challenging. To narrow the scope of this paper, we focus on extensions of SINDy in this setting. Broadly, existing approaches fall into two categories: simultaneous and two-step. Simultaneous methods perform denoising and system identification jointly \cite{wentz2023derivative,wang2023simultaneous}. Some explicitly model noise as an additional variable and optimize it alongside the SINDy coefficient matrix, allowing the identification process to help estimate the noise \cite{rudy2019deep,kaheman2022automatic}. Others couple SINDy with Runge–Kutta schemes \cite{goyal2022discovery,forootani2023robust,zhai2023parameter,zhai2024state}, which link the noise variables and SINDy coefficients through discretized dynamics, enabling joint optimization.

By contrast, two-step methods first denoise the state variables and estimate their first-order derivatives, then feed these estimates into SINDy \cite{lubansky2006general,kostelich1993noise,kasac2018algebraic}. Classical instances include local-regression smoothers \cite{green_nonparametric_1993} such as the Savitzky–Golay filter \cite{1964Smoothing,schafer2011savitzky,ahnert2007numerical}. Regularization-based formulations impose smoothness-promoting penalties \cite{stickel_data_2010}, including smoothing splines \cite{sun_physics-informed_2021,1975Smoothing} and total variation regularization \cite{tikhonov1963regularization,vogel2002computational,chartrand2011numerical}. More recently, deep learning–based denoisers have also been explored \cite{zhu2019seismic,fan2020vibration,yu2019deep}.

Simultaneous approaches offer end-to-end identification directly from noisy data, but they typically incur higher model complexity and computational cost, particularly in high-dimensional settings, and often require careful tuning to remain stable. Two-step pipelines are generally more modular and computationally lighter; they are also agnostic to the downstream identification method, making them easy to pair with SINDy or alternative data-driven models. Despite these advantages, recent work has largely emphasized simultaneous formulations, and robust, scalable two-step strategies remain comparatively underexplored.

Motivated by these limitations, we propose a two-step approach for modeling nonlinear dynamical systems: RKTV-INR (Runge-Kutta and Total Variation Based Implicit Neural Representation) built on implicit neural representations (INRs) \cite{park2019deepsdf,mildenhall2021nerf}. We represent the system trajectory as a continuous function with an INR model fitted to noisy observations, and impose constraints during training to steer the fit away from measurement noise. The INR output is treated as denoised state data, while first-order derivatives are obtained via automatic differentiation \cite{baydin2018automatic}. These denoised states and derivatives are then supplied to SINDy to identify the governing dynamics. Suppose the dynamics of $x \in \mathbb{R}^n$ are generated by the ODE $\frac{\mathrm{d}x}{\mathrm{d}t} = f(x)$, and define $g(t) = f(x(t))$. Our main contributions are:

\begin{enumerate}
    \item We apply Runge--Kutta integration within the denoising stage to recover $(x(t), g(t))$ consistent with the measurements; by contrast, prior work typically uses Runge--Kutta schemes to learn $f(x)$ directly.
    \item We enforce three complementary constraints to (i) match state values at observation times, (ii) match time derivatives at those times, and (iii) promote smoothness of the reconstructed state and derivative trajectories.
    \item We cast denoising as learning $g:\mathbb{R}\to\mathbb{R}^n$ over the data interval $[0,T]$ (scalar input $t$), avoiding assumptions about generalization outside this window. In contrast, approaches that learn $f:\mathbb{R}^n\to\mathbb{R}^n$ from noisy state samples \cite{rudy2019deep,kaheman2022automatic,goyal2022discovery} must generalize from off-trajectory observations, which is particularly challenging when noise displaces states from the true manifold.
\end{enumerate}

The paper is organised as follows. \autoref{sect2} summarises the SINDy algorithm. \autoref{sect3} reviews the most effective two-step approaches. \autoref{sect4} introduces RKTV-INR, our two-step INR-based framework for data-driven modeling with time dependence and noise. \autoref{sect5} presents experiments demonstrating the effectiveness of the proposed method. Finally, \autoref{sect6} concludes and outlines directions for future work.

\section{Overview of the SINDy Algorithm}
\label{sect2}

%The sparse identification of nonlinear dynamics (SINDy) algorithm identifies the governing equation of the target system from data. Specifically, the required inputs are the system's state and first-order derivative data. Suppose the system contains variables $x_1,\cdots,x_n$, and we monitor the system at time points $t_1,\cdots, t_m$. Then, the state and derivative data can be seen as two matrices,

The sparse identification of nonlinear dynamics (SINDy) algorithm infers a system’s governing equations directly from data. Its inputs are time series of the system state and the corresponding first-order derivatives. Suppose the system has variables $x_1,\ldots,x_n$ observed at times $t_1,\ldots,t_m$; then these measurements naturally assemble into two $m\times n$ matrices, one containing the state samples and the other their time derivatives

\begin{align}
    \mathbf{X} =  \begin{bmatrix}
    x_1(t_1) &\cdots & x_n(t_1) \\
    \vdots & \ddots & \vdots \\
    x_1(t_m)  & \cdots & x_n(t_m)
\end{bmatrix},     \dot{\mathbf{X}} = \begin{bmatrix}
    \dot{x}_1(t_1)  & \cdots & \dot{x}_n(t_1) \\
    \vdots  & \ddots & \vdots \\
    \dot{x}_1(t_m) & \cdots & \dot{x}_n(t_m)
\end{bmatrix},
\end{align}

%\noindent where $\dot{x}_i(t_j) = \frac{\mathrm{d} x_i}{\mathrm{d}t} |_{t_j}$. Each column of the state matrix $\mathbf{X}$ represents the trajectory of one variable in the system. This matrix is measured by sensors in the real situation, so measurement errors are unavoidable. In addition, the acquisition of the derivative matrix $\dot{\mathbf{X}}$ is much more difficult. Generally, it is not available from real-world sensors and needs to be approximated by numerical methods, such as the central difference \cite{epperson2013introduction}. However, this approximation is not robust and will amplify the measurement errors in the state matrix, which will be discussed in \autoref{sect2.4}.

\noindent where $\dot{x}_i(t_j) = \frac{\mathrm{d} x_i}{\mathrm{d}t} |_{t_j}$.
Each column of the state matrix $\mathbf{X}$ records the time trajectory of a single state variable. In practice, $\mathbf{X}$ is obtained from sensors and therefore inevitably contains measurement noise. Estimating the derivative matrix $\dot{\mathbf{X}}$ is even more challenging: sensors rarely provide derivatives directly, so $\dot{\mathbf{X}}$ is typically computed from $\mathbf{X}$ via numerical differentiation (e.g., central differences \cite{epperson2013introduction}). Unfortunately, such approximations are prone to amplifying measurement noise and are thus fragile.

%For this system, the governing equation, or dynamics, is defined as a function $\mathbf{f} = [f_1,\cdots,f_n]: \mathbb{R}^{n} \rightarrow \mathbb{R}^{n}$ such that

%\begin{align}
%    \dot{\mathbf{X}} = \mathbf{f}(\mathbf{X}). \label{fun3.5}
%\end{align}

%The function $\mathbf{f}$ is often nonlinear, which makes it difficult to compute directly. SINDy assumes that the governing equation of each individual variable $f_i$ can be seen as a linear combination of known basis functions. Hence, we need to select some basis functions to span a function space. In SINDy, the set containing all basis functions is called `\textit{function library}' and denoted as $\Theta  = \{ \theta_1, \cdots,\theta_k | \theta_i : \mathbb{R}^n \rightarrow \mathbb{R}, \ \forall i = 1,\cdots, k \}$. Hence, $f_i = \sum_{j=1}^k \xi_{ji} \cdot \theta_j , \ \forall i = 1 ,\cdots,n $. 

For this system, the dynamics are given by a function $f = [f_1,\ldots,f_n]: \mathbb{R}^n \to \mathbb{R}^n$ such that
\begin{align}
\dot{\mathbf{X}} = f(\mathbf{X}), \label{fun3.5}
\end{align}
interpreted row-wise over the sampled states. The function $f$ is generally nonlinear and difficult to compute directly. SINDy posits that each component $f_i$ can be expressed as a sparse linear combination of known basis functions. We therefore choose a collection of candidate functions, called the \emph{function library}, $\Theta = \{\theta_1,\ldots,\theta_k \mid \theta_j:\mathbb{R}^n \to \mathbb{R}\}$, and write
$f_i(x) \approx \sum_{j=1}^k \theta_j(x) \xi_{ji}$ for $i=1,\ldots,n$, with coefficients $\xi_{ji}$ that are predominantly zero. The matrix form of the above formulas is

\begin{align}
    \begin{bmatrix}
        f_1 & \cdots & f_n
    \end{bmatrix} = \begin{bmatrix}
        \theta_1 & \cdots & \theta_k
    \end{bmatrix} \times \begin{bmatrix}
        \xi_{11} & \cdots & \xi_{1n} \\
        \vdots & \ddots & \vdots \\
        \xi_{k1} & \cdots & \xi_{kn} 
    \end{bmatrix}. 
    \label{fun3.7}
\end{align}

Notice that $f_i$ represents the derivative of each single variable $x_i$, i.e. $ f_i(\boldsymbol{x}(t))=  \dot{x}_i(t)$. Hence, the discrete version of Equation \eqref{fun3.7} is

\begin{align}
    \dot{\mathbf{X}} = \Theta \times \Xi, \label{fun3.11}
\end{align}

\noindent where

\begin{align}
    \Xi = \begin{bmatrix}
        \xi_{11} & \cdots & \xi_{1n} \\
        \vdots & \ddots & \vdots \\
        \xi_{k1} & \cdots & \xi_{kn} 
    \end{bmatrix}.    
\end{align}

%The choice of $\Theta$ often relies on the state data $\mathbf{X}$, so $\Theta = \Theta(\mathbf{X})$. We take a system with two variables $(x_1,x_2)$ as an example. The function library may contain constant, polynomial, and trigonometric functions, and its choice is very variable: 

The choice of the function library $\Theta$ is typically data-dependent, so we write $\Theta=\Theta(\mathbf{X})$. As an example, for a two-variable system $(x_1,x_2)$, a common library includes constants, polynomials, cross terms, and trigonometric functions

\begin{align}
    \Theta(\mathbf{X}) =\{
        \boldsymbol{1}, \mathbf{X}^{p_1}, \mathbf{X}^{p_2}, \cdots, \mathbf{X}^{p_i}, \cdots, \sin{\mathbf{X}}, \cos{\mathbf{X}},\cdots
    \}
\end{align}

\noindent where $\mathbf{X}^{p_i}$ represents the $i$-th order polynomial function library. For example, the second-order polynomial function library is given by

\begin{align}
    \mathbf{X}^{p_2} = \begin{bmatrix}
        x_1^2(t_1) & x_2^2(t_1) & (x_1x_2)(t_1) \\
        \vdots & \vdots & \vdots \\
        x_1^2(t_m) & x_2^2(t_m) & (x_1x_2)(t_m) 
    \end{bmatrix}.
\end{align}

%To determine the coefficient matrix $\Xi$, the final step is to perform regression on \eqref{fun3.11}. Many physical systems only require a few terms to define the dynamic behaviour, which means that even though we create a high-dimensional function space, our final selected function is sparse in this space \cite{kwapien2012physical,wang2011predicting,bertsimas2020sparse}. To reach this solution, we need to perform a sparse regression in Equation \eqref{fun3.11}. Commonly used methods for this step include Lasso regression \cite{10.1111/j.2517-6161.1996.tb02080.x} and sequential thresholded least-squares \cite{brunton2016discovering}. The identified coefficient matrix provides an estimation for the governing equation.

To estimate the coefficient matrix $\Xi$, we regress the relation in \eqref{fun3.11}. As many physical systems are parsimonious, only a few terms govern the dynamics, the true model is sparse within the (potentially high-dimensional) function library \cite{kwapien2012physical}. We therefore solve a sparse regression problem for \eqref{fun3.11}. Common choices include $\ell_1$-regularised least squares (Lasso) \cite{10.1111/j.2517-6161.1996.tb02080.x} and sequentially thresholded least squares \cite{brunton2016discovering}. The resulting $\Xi$ specifies the estimated governing equations.

%The accuracy of the SINDy algorithm has been well established when applied to perfect data. However, as previously mentioned, its performance is often limited by two key issues. First, the state data measured are often noisy, leading to inaccuracy of the function library. Second, it requires data on the first-order derivative, which is impractical in real scenarios. If it is approximated by numerical methods, the noise in the state data will be amplified. Consequently, this paper proposes a data denoising method that can obtain denoised data as well as first-order derivative data. Then, combined with the SINDy algorithm, the governing equation of the dynamic system is derived.

The accuracy of SINDy is well established on clean, noise-free data. In practice, however, performance is hindered by two issues. First, sensor noise corrupts the state measurements, degrading evaluations of the function library $\Theta(\mathbf{X})$. Second, SINDy requires first-order derivatives, which are rarely measured directly; when estimated numerically, these derivatives tend to amplify noise \cite{chartrand2011numerical}. To address both challenges, we introduce a denoising procedure that reconstructs the state trajectory and its first derivative from noisy observations. The resulting denoised states and derivatives are then supplied to SINDy to recover the governing equations of the dynamical system.

\section{Related work on Two-Step Approaches}
\label{sect3}

%This paper focuses on denoising methods for the two-stage approach to modeling dynamic systems. So we review several representative approaches in this category.

This paper focuses on denoising via the two-stage approach for modeling dynamical systems. Accordingly, we briefly review representative methods in this class, highlighting how they suppress measurement noise and produce state and derivative estimates for subsequent identification.

\subsection{The Savitzky-Golay Filter}

%The Savitzky-Golay filter, abbreviated as S-G filter, was originally proposed in \cite{1964Smoothing}. It can identify the noise while still maintaining the trend, period and other key information of the signal.

%The S-G filter has two parameters: window length $l = 2k + 1, k \in \mathbb{N}^+$ and polynomial function order $s-1 \in \mathbb{N}^+$. At each time point $t_i$, a sliding window $(\mathbf{X}_{i-k},\cdots,\mathbf{X}_{i},\cdots,\mathbf{X}_{i+k})$ is created, where the subscript represents the row number. Then, a polynomial function of order $(k-1)$ is constructed to perform regression on the data in the window. The polynomial function then provides an estimation of the denoising state data, and the first-order derivative is approximated by the derivative of the polynomial function.

%In the process of reconstructing Fourier transform infrared spectra, Zhao et al. \cite{Zhao2014} chose S-G filters to smooth and denoise the spectra.

The Savitzky–Golay (S–G) filter, introduced in \cite{1964Smoothing}, smooths noisy measurements while preserving local structure such as trends, peaks, and periodicity. It has two hyperparameters: the window length $l=2k+1$ with $k\in\mathbb{N}^+$, and the polynomial degree $s-1$ with $s\in\mathbb{N}^+$. For each time index $i$, a sliding window $(\mathbf{X}_{i-k},\ldots,\mathbf{X}_i,\ldots,\mathbf{X}_{i+k})$ is formed (rows of $\mathbf{X}$). A degree-$(s-1)$ polynomial is fit by least squares to the samples in the window; evaluating this polynomial at the center yields the denoised state at $t_i$, and differentiating it provides an estimate of the first-order derivative. In practice, each state variable (each column of $\mathbf{X}$) is filtered independently.

\subsection{The Smoothing Spline Method}

%The smoothing spline method was first proposed by Reinsch in 1975 \cite{1975Smoothing}. The core idea is to handle noisy data by balancing the accuracy of data fitting and the smoothness of the function. For given noisy data, the aim is to find a smooth function $f(x)$ that can not only fit the data well (to reduce bias) but also avoid overfitting to noise (to control variance). The optimal solution $f(x)$ has been proven to be a natural cubic spline.

%Silverman \cite{Silverman1984A} proposed an approximate cross validation method and discussed the cross validation method for smooth parameter selection in spline regression. Shang and Cheng \cite{2012Local} studied local and global inference for smooth spline estimation in a unified asymptotic framework, introduced the functional Bahadur representation as a new technical tool, and developed four interrelated inference programs. Ahnert and Abel \cite{ahnert2007numerical} systematically compared the performance differences between local methods (such as S-G filter, finite difference) and global methods (such as smoothing spline).

The smoothing‐spline approach, introduced by Reinsch \cite{1975Smoothing}, treats denoising as a balance between data fidelity and smoothness. Given noisy observations, it estimates a function $f(x)$ by minimizing a penalized least-squares objective—typically the sum of squared residuals plus a roughness penalty on curvature (e.g., $\int (f''(x))^2\mathrm{d}x$). The solution is a natural cubic spline with knots at the data, which achieves a bias–variance trade-off by tuning a single smoothing parameter. For selecting this parameter, Silverman \cite{Silverman1984A} proposed an approximate cross-validation criterion and discussed cross-validation strategies for spline regression. Shang and Cheng \cite{2012Local} developed a unified asymptotic framework for local and global inference with smoothing splines, introducing a functional Bahadur representation and related inference procedures.

\subsection{Total Variation Regularisation}

%The solution of total variation regularisation is derived by finding the minimiser of the Tikhonov-TV function \cite{tikhonov1963regularization}

%\begin{align}
%    T(u) = \int_a^b \left( K(u) - \mathbf{X} \right)^2 \ \mathrm{d}t + \alpha \int_a^b \left| \frac{\mathrm{d} u}{\mathrm{d} t}\right|\  \mathrm{d}t. \label{fun4.10}
%\end{align}

%\noindent where $K(u)$ is a function operator, and $\mathbf{X}$ is regarded as a noisy measurement of the state variable. The term on the right-hand side is total variation, which can be seen as a metric to measure the smoothness of the function $u$. $\alpha$ is a small positive coefficient. The solution satisfies $K(u) \approx \mathbf{X}$ while still maintaining smoothness and thus is noise-free. When $K(u) \equiv u$, $u$ provides an estimation of the state data; while when $K$ is the integration operator, $u$ provides an estimation of the first-order derivative.

%In the discretised setting, $u$ is a vector $\in \mathbb{R}^n$ and the optimisation step is based on calculating the gradient and performing gradient descent \cite{vogel2002computational,chartrand2011numerical}.

Total variation (TV) regularisation estimates a clean signal $u$ by minimising the Tikhonov-TV functional
\begin{equation}
T(u) = \int_a^b \big(K(u)-\mathbf{X}\big)^2\,\mathrm{d}t \;+\; \alpha \int_a^b \left|\frac{\mathrm{d}u}{\mathrm{d}t}\right|\,\mathrm{d}t.
\label{fun4.10}
\end{equation}
Here, $K(u)$ is a forward/operator map and $\mathbf{X}$ denotes noisy measurements of the state. The first term enforces data fidelity, while the second is the TV penalty, which measures the total variation of $u$ and promotes piecewise-smooth trajectories by discouraging spurious oscillations. The parameter $\alpha>0$ controls the trade-off between fidelity and regularity. The minimiser of \eqref{fun4.10} satisfies $K(u)\approx \mathbf{X}$ while avoiding overfitting noise. In particular, when $K\equiv I$, $u$ estimates the state; when $K$ is the integration operator, $u$ estimates the first derivative (since its integral matches the data). In the discretised setting, $u\in\mathbb{R}^n$ at sampled times and the optimisation is typically solved with gradient-based iterations (e.g., subgradient or proximal methods) \cite{vogel2002computational,chartrand2011numerical}.

\subsection{Implicit Neural Representation Based Approach}
\label{sect2.4}

Recent advances in deep learning \cite{bishop2023deep, ramasinghe2023effectiveness, saratchandran2024sampling} have enabled practical, complex data-driven models. Within this paradigm, implicit neural representations (INRs) encode signals and data as continuous functions \cite{park2019deepsdf,mildenhall2021nerf, saratchandran2023curvature, saratchandran2024activation}, typically using multilayer perceptrons (MLPs) with tailored activations such as sinusoidal \cite{sitzmann2020implicit}, wavelets \cite{saragadam2023wire} or sinc functions \cite{saratchandran2024sampling}. Among them, Sitzmann et al. \cite{sitzmann2020implicit} proposed a unique parameter initialization strategy for MLPs with sinusoidal activation functions to enable efficient learning and feature extraction, and named this model Sinusoidal Representation Networks or SIREN. SIREN-based implicit neural representations have shown strong denoising behaviour in practice: Saitta et al. \cite{saitta2024implicit} used it to denoise and super-resolve 4D-flow MRI velocity fields in the thoracic aorta, while Kim et al.  \cite{kim2022zero} proposed a zero-shot INR denoiser that constrains weight growth to exploit INRs’ implicit priors.

Similarly, in this work, we apply the SIREN-based INRs to perform data denoising in the field of dynamic systems. Specifically, we model the trajectory as a time-conditioned INR using SIREN, representing the mapping $t \mapsto (x_1,\ldots,x_n)$ with a MLP that utilises a sine function as its activation, $\boldsymbol{\chi}_{\boldsymbol{\theta}}(t)$. We fit $\boldsymbol{\chi}_{\boldsymbol{\theta}}(t)$ to measurements $\{(t_i,\mathbf{X}_{i,:})\}_{i=1}^m$ by minimising a data-fidelity objective with weight decay:
\begin{equation}
    \mathcal{L}(\boldsymbol{\theta})
    = \frac{1}{m}\sum_{i=1}^{m}
    \left\|
        \boldsymbol{\chi}_{\boldsymbol{\theta}}(t_i) - \mathbf{X}_{i,:}
    \right\|_2^2
    + \lambda \left\|\boldsymbol{\theta}\right\|_2^2.
\end{equation}
The second term penalty mitigates overfitting and encourages smooth reconstructions, yielding effectively denoised state estimates. First-order time derivatives are then obtained directly via automatic differentiation of $\boldsymbol{\chi}_{\boldsymbol{\theta}}(t)$.

\section{Main Contribution: RKTV-INR}
\label{sect4}

%This paper proposes a new two-stage method for modeling dynamical systems. Its main innovation is called RKTV-INR, which aims to denoise data and estimate its first-order derivative. 

This section introduces a novel two-step framework for modeling dynamical systems. The core innovation, RKTV-INR (Runge-Kutta and Total Variation Based Implicit Neural Representation), provides a methodology to robustly denoise data and produce accurate estimates of first-order derivatives.

%\subsection{Formulation}

%Runge-Kutta integration and total variation regularizer are employed in the method. The input of the algorithm is a vector of time points $t = (t_1, \cdots,t_m)$ and the noisy measurement of the state variables at these time points, denoted as matrix $\mathbf{X}$. It is also assumed that the step length $h$ between time points is a constant. The outcome of the algorithm is the estimation of the state and first-order derivative data, which should meet three criteria:

\subsection{Formulation}

The goal of this framework is to perform state and first-order derivative estimation from noisy real-world measurements, where the resulting estimates can then be utilized for data-driven dynamical system modeling. The inputs are a vector of time points $t=(t_1,\ldots,t_m)$, and noisy measurements of the state at these times, assembled in the matrix $\mathbf{X}$. Without loss of generality, we may assume that the length between two time points is a constant $h$.

\subsection{Architecture}
\label{section4.2}

%According to the three requirements demonstrated above, we correspondingly set up three loss functions in the RKTV-INR.

Our framework, RKTV-INR, is based on the INR approach introduced in \autoref{sect2.4}. More concretely, we employ SIREN as our base INR to represent the continuous function from time $t$ to the state variables in the system $(x_1,\ldots,x_n)$, denoted as $\boldsymbol{\chi}_{\boldsymbol{\theta}}(t): \mathbb{R} \rightarrow \mathbb{R}^n,\ t \mapsto (x_1,\ldots,x_n)$ and $\boldsymbol{\theta}$ represents the set of parameters in the INR (SIREN). The other hyperparameters of the model are variable as needed. In \autoref{sect5.1}, we provide a detailed specification of the hyperparameters used in the experiments. The INR model is trained on the measured state dataset $\boldsymbol{X}$, and then the trained INR is used to recover denoised estimates of the state and its first-order time derivative. We require that the estimates satisfy the following three criteria:

\begin{enumerate}
    \item The output of INR at observed time points should match the ground-truth state values.
    \item The derivative of INR, calculated by automatic differentiation, at observed time points should match the ground-truth derivative values, and the training process does not rely on access to derivative data.
    \item The smoothness of the estimated state and its derivative curves should be guaranteed to remain consistent with the behavior of real physical systems.
\end{enumerate}

To enforce the above three criteria, we introduce three corresponding loss terms that are used to train our INR model, RKTV-INR.

\subsubsection*{State Fitting}

First, to enable state fitting, the output of the INR $\boldsymbol{\chi}_{\boldsymbol{\theta}}(t)$ should match the state data $\mathbf{X}$. Hence, the first loss function is 

\begin{align}
     \mathcal{L}_1(\boldsymbol{\theta}) = \frac{1}{m} \sum_{i=1}^m \left|\boldsymbol{\chi}_{\boldsymbol{\theta}}(t_i) - \mathbf{X}_{i,\cdot} \right|_2^2.
\end{align}

This loss function is used to train the INR to represent the state variables as continuous functions, thereby capturing the temporal features and patterns of the state variables. However, the INR inevitably also learns noise from the measurements $\mathbf{X}$, which can introduce errors in state estimates.

\subsubsection*{Runge-Kutta Integration}

%The second condition is more complex, but it is also the main innovation of this algorithm. As a recap, we first review the Runge-Kutta integration, which is a famous method for performing numerical integration \cite{epperson2013introduction}. We take the 4-th order Runge-Kutta integration as the example. Consider a function $\boldsymbol{x}(t): \ \mathbb{R} \rightarrow \mathbb{R}^n$ that is governed by the ODE below

The second loss function enhances the INR’s ability to learn and estimate the temporal derivatives of the state variables through Runge-Kutta integration. As a recap, Runge-Kutta integration is a numerical method employed to approximate the integral of governing equations $\boldsymbol{f}(\mathbf{X},t)$ of ordinary differential equations (ODEs) over intervals \cite{epperson2013introduction}. Given initial conditions, it is frequently used to simulate the solution of ODEs. The 4th order Runge-Kutta scheme is most commonly used in practice.

When the second criterion requires the INR derivatives to match the ground-truth derivatives of state variables , we implicitly establish the following ODE:
\begin{align}
    \frac{\mathrm{d} \mathbf{X}}{\mathrm{d} t} = \boldsymbol{f}(\mathbf{X},t) = \frac{\mathrm{d} \mathbf{\chi}_{\boldsymbol{\theta}}}{\mathrm{d} t}(t).
\end{align}
We notice that the governing equation $\frac{\mathrm{d} \mathbf{\chi}_{\boldsymbol{\theta}}}{\mathrm{d} t}$ is state independent and can be directly computed by automatic differentiation. To integrate this equation using the 4th Runge-Kutta scheme, we first define the following four terms:

\begin{align}
\begin{cases}
    k_1(\mathbf{X},t) = h \cdot \boldsymbol{f}(\mathbf{X},t) = h \cdot \frac{\mathrm{d} \mathbf{\chi}_{\boldsymbol{\theta}}}{\mathrm{d} t}(t) \\
    k_2(\mathbf{X},t) = h \cdot \boldsymbol{f}(\mathbf{X}+\frac{k_1}{2},t+ \frac{h}{2}) = h \cdot \frac{\mathrm{d} \mathbf{\chi}_{\boldsymbol{\theta}}}{\mathrm{d} t}(t + \frac{h}{2})   \\
    k_3(\mathbf{X},t) = h \cdot \boldsymbol{f}(\mathbf{X}+\frac{k_2}{2},t+ \frac{h}{2})= h \cdot \frac{\mathrm{d} \mathbf{\chi}_{\boldsymbol{\theta}}}{\mathrm{d} t}(t + \frac{h}{2})  \\
    k_4(\mathbf{X},t) = h \cdot \boldsymbol{f}(\mathbf{X}+k_3,t+h)  = h \cdot \frac{\mathrm{d} \mathbf{\chi}_{\boldsymbol{\theta}}}{\mathrm{d} t}(t+h) 
\end{cases}.
\end{align}
We then define the 4th order Runge-Kutta operator as $\mathrm{RK}(\mathbf{X},t, \frac{\mathrm{d} \mathbf{\chi}_{\boldsymbol{
\theta}}}{\mathrm{d} t}, h) = \frac{1}{6}[k_1 + 2k_2 + 2k_3 + k_4](\mathbf{X},t)$. 

Although the Runge-Kutta operator is typically used to solve $\mathbf{X}$, in our work, the solution of the state variable $\mathbf{X}$ is known, while the INR needs to be trained. We achieve the training by optimizing the following loss function:

\begin{align}
\mathcal{L}_2(\boldsymbol{\theta}) = \frac{1}{m-1} \sum_{i=1}^{m-1} \left| (\mathbf{X}_{i+1} - \mathbf{X}_{i}) - \mathrm{RK} (\mathbf{X}_{i} ,t_i, \frac{\mathrm{d}\mathbf{\chi}_{\boldsymbol{\theta}}}{\mathrm{d} t},h))\right|_2^2. \label{fun4.18}
\end{align}
More directly, we aim to solve the inverse problem of the Runge-Kutta scheme so that the INR derivatives can be optimised to match the ground-truth derivatives using the state data. Compared with standard INR training, this loss function greatly enhances the INR’s ability to estimate derivatives. Traditional INR models may converge to the state data in their outputs, but their estimated derivatives cannot be guaranteed to converge to the true derivatives. This issue has also been mentioned in Sobolev training \cite{10.5555/3294996.3295182}.

\begin{figure}[t]
    \centering
    \includegraphics[width=1\linewidth]{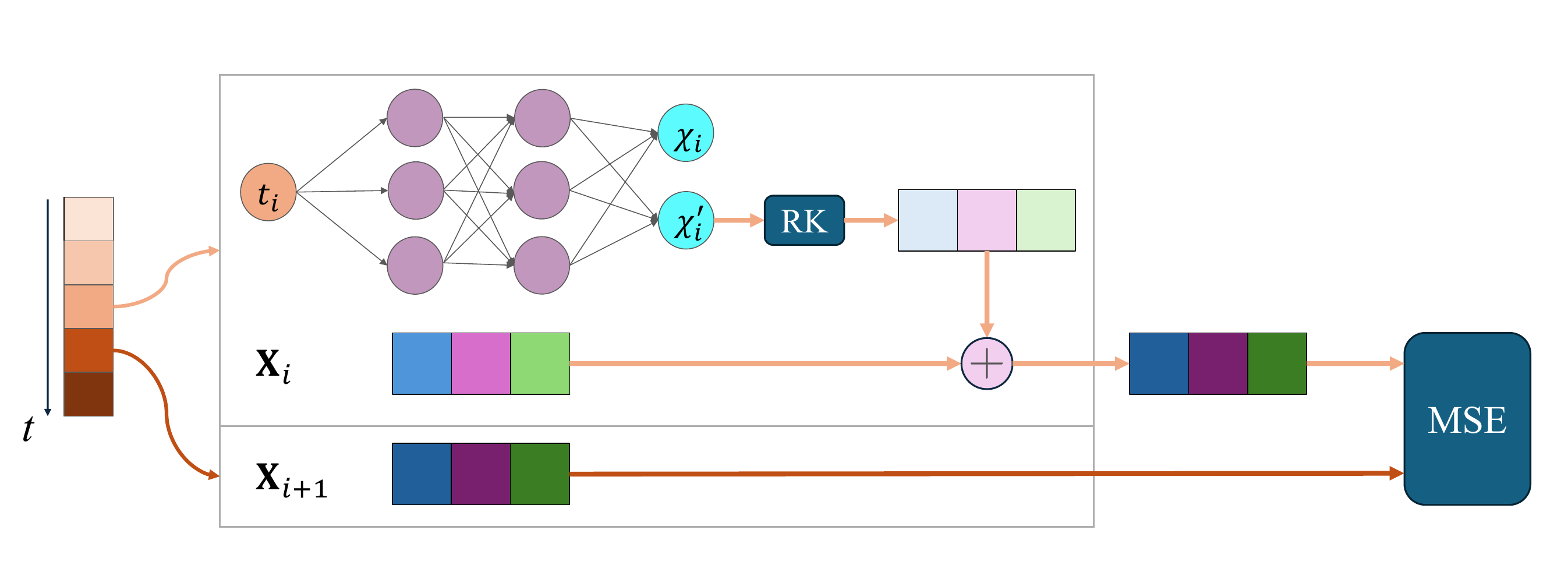}
    \caption{Diagram of the Runge-Kutta loss function. If the INR can accurately estimate the first-order derivative, then its Runge-Kutta integration over a time interval can compensate for the discrepancy in the state between two time points.}
    \label{fig5.7}
\end{figure}

To provide a theoretical interpretation of its working principle, we first suppose that there is a smooth and differentiable function $\mathbf{X}(t)$. Then the difference between the function value at two nearby time points should be equal to the Runge-Kutta integration of the derivative of $\mathbf{X}(t)$. This expression is the fundamental design principle of the Runge-Kutta scheme. Hence, if the difference term and the Runge-Kutta operator in Equation \eqref{fun4.18} are equal to each other, the derivative of INR $\frac{\mathrm{d}\mathbf{\chi}_{\boldsymbol{\theta}}}{\mathrm{d} t}$ is converging to the derivative of the target function $\frac{\mathrm{d} \mathbf{X}(t)}{\mathrm{d} t}$. In all, this loss function is an implicit way for INR to learn derivative information from state data $\mathbf{X}$. Note that this process does not require derivative data. Figure \ref{fig5.7} shows a schematic illustration of how it works.

\subsubsection*{Smoothness}

%To meet the third requirement, we have to design a more efficient regularizer that forces the state and derivative curves to be smooth. One fact is that if the derivative of a function is smooth, then the function itself will also be smooth, and its smoothness is higher than its derivative. Hence, if the second-order derivative of the neural network is smooth, then the function and the first-order derivative are guaranteed to be smooth. Consequently, we set up the third loss function as follows:

To satisfy the third criterion, we introduce a regularizer that promotes smooth state and derivative trajectories. Intuitively, if a function’s derivative is smooth, then the function is at least as smooth; in particular, enforcing smoothness of the second derivative ensures smoothness of both the function and its first derivative. Accordingly, we penalise second-order variations of INR output, leading to the following loss term:

\begin{align}
\mathcal{L}_3 =  \frac{1}{m-1} \sum_{i=1}^{m-1} \left| \frac{\mathrm{d}^2 \mathbf{\chi}_{\boldsymbol{\theta}}}{\mathrm{d} t^2}(t_{i+1}) - \frac{\mathrm{d}^2\mathbf{\chi}_{\boldsymbol{\theta}}}{\mathrm{d} t^2}(t_{i})\right|_2^2. \label{fun4.23}
\end{align}

%By forcing the difference between the second-order derivative at two nearby time points to be small, the second-order derivative can be a smooth curve, thus making our estimation be smooth curve as well. In real physical systems, continuous variables exhibit smooth behaviour over local time intervals, even in chaotic systems. Therefore, this condition can help us obtain more accurate estimations.

By penalising variations in the second derivative across neighbouring time points, we enforce a smooth second-derivative trajectory, which in turn yields smooth estimates of both the state and its first derivative. Continuous variables in physical systems typically evolve smoothly over local time intervals, even in chaotic regimes, so this regularity prior improves estimation accuracy.

%It is worth noting that the loss function represented by Equation \eqref{fun4.23} can be interpreted as the total variation of the second-order derivative. In the limit that the time step $h$ approaches 0, it is identical to

%\begin{align}
%    \mathcal{L}_3 = \frac{h}{m-1} \int_a^b \left[ \frac{\mathrm{d}}{\mathrm{d} t} \left(\frac{\mathrm{d^2 \chi_{\boldsymbol{\theta}}}}{\mathrm{d} t^2} \right) 
%\right]^2 \mathrm{d} t.
%\end{align}
%Therefore, we adopt a similar idea to total variation regularisation here and extend it to higher-order derivatives. The derivation of this identity is in the Appendix \ref{A1}.

It is worth noting that the loss in \eqref{fun4.23} can be interpreted as a discrete total variation penalty on the second time derivative. In the limit as the step size $h \to 0$, it converges to
\begin{align}
    \mathcal{L}_3
    = \frac{h}{m-1} \int_a^b
    \left[
        \frac{\mathrm{d}}{\mathrm{d}t}
        \!\left(\frac{\mathrm{d}^2 \chi_{\boldsymbol{\theta}}}{\mathrm{d} t^2}\right)
    \right]^2
    \mathrm{d} t .
\end{align}
Thus, we adopt a TV-style regularisation, here applied to higher-order derivatives. The derivation of this identity is provided in Appendix~\ref{A1}.

\subsubsection*{A New Training Loss Function}

%In all, the RKTV-INR can be seen as a combination of INR, Runge-Kutta integration and total variation regularisation. The loss function used in the RKTV-INR is a linear combination of the three individual ones:

%\begin{align}
%    \mathcal{L} = c_1 \cdot  \mathcal{L}_1 + c_2  \cdot \mathcal{L}_2 + c_3 \cdot \mathcal{L}_3, \ \ \ c_1,c_2,c_3>0.
%\end{align}
%As a rule of thumb, in implementations, we normally choose $c_1 = c_2 = 1$. For the smoothness coefficient $c_3$, its choice is variable and dependent on the noise level of the data and the smoothness of the system trajectory. The interval of the choice can be $[10^{-5},1]$, and $10^{-2}$ is the most common choice.

In summary, RKTV-INR loss combines an implicit neural representation model, Runge-Kutta integration, and total variation regularisation. The training objective is a weighted sum of three terms:
\[
\mathcal{L} = c_1\,\mathcal{L}_1 + c_2\,\mathcal{L}_2 + c_3\,\mathcal{L}_3, \qquad c_1,c_2,c_3>0.
\]
In practice, we typically set $c_1=c_2=1$. The smoothness weight $c_3$ is tuned to the noise level and expected trajectory regularity; values in $[10^{-5},1]$ are effective, with $10^{-2}$ a common default. A larger $c_3$ encourages greater smoothness in the results.

%Figure \ref{fig5.8} shows the overall diagram of the RKTV-INR, which combines implicit neural representation, Runge-Kutta integration, and total variational regularisation to estimate the true system state as well as its first-order derivatives. Our data consists of grid points along a time axis and noisy observations of the system variables' states at the corresponding time points. During the model training, we use a neural network to represent the system variables and to output the state, first-order, and second-order derivatives. We define three loss functions and use them to update the parameters of the neural network. After training, the neural network is used to estimate the noise-free system states and their first-order derivatives.

\begin{figure}[t]
    \centering
    \includegraphics[width=1\linewidth]{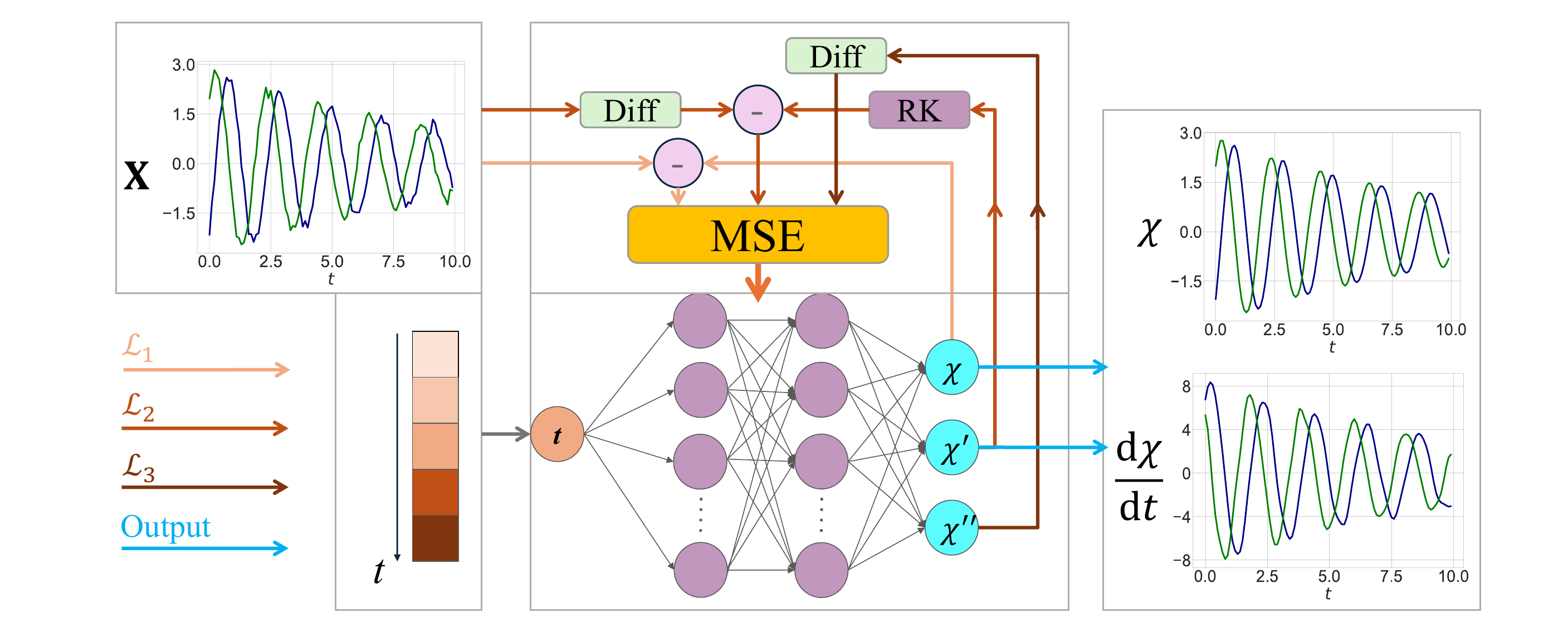}
    \caption{The overall diagram of the RKTV-INR. A SIREN-based INR is used to represent the state variables $\chi(t)$. RKTV-INR combines state fitting with total variation and the Runge-Kutta integration scheme. After training, it provides an estimation of the state and first-order derivative.}
    \label{fig5.8}
\end{figure}

Figure \ref{fig5.8} illustrates the Runge–Kutta training pipeline, which combines an implicit neural representation, Runge–Kutta integration, and total variation regularisation to recover the true system state and its first-order derivatives. The data consist of uniformly spaced time samples and noisy state measurements at those times. During training, the INR parameterises the trajectory and produces the state together with its first and second-order time derivatives (via automatic differentiation). Three loss terms, data fidelity, Runge–Kutta integration, and second-derivative smoothness, are used to update the INR parameters. After optimisation, the INR yields denoised state estimates and accurate first-order derivatives.

\section{Experiments}
\label{sect5}

This section demonstrates the performance of RKTV-INR by through various experiments.

\subsection{Experimental design}
\label{sect5.1}

%To answer these questions, we simulate the state data of actual dynamic systems, with artificial noise added with variance $\zeta^2$. We first select a real existing dynamical system and obtain the real ground-truth state data. Then we define the average length scale as $L = \sqrt{\frac{1}{n}\sum_{j=1}^n\mathrm{Var}(\mathbf{X}_{\cdot,j})}$. Each column of $\mathbf{X}$ represents the trajectory of one variable. Hence, the average length scale is obtained by averaging the variances of all variables in the system and then take the root squre. After that, noise satisfying a specific distribution is added to the original data. In this paper, we mainly used Gaussian noise to process the data. Suppose that the real ground-truth state data is $\mathbf{X}$. The intensity of noise can be controlled by adjusting a predetermined parameter - relative noise level $\sigma^2$, which is defined as $\sigma^2 = \frac{\zeta^2}{L^2}$This method can adaptively control the scale of the added noise according to the scale of the system. 

We simulate state trajectories of real dynamical systems and corrupt them with additive noise of variance $\zeta^2$. For a chosen system, we first generate the ground-truth state matrix $\mathbf{X}$ then define an average length scale
\[
L = \sqrt{\frac{1}{n}\sum_{j=1}^n \operatorname{Var}\!\big(\mathbf{X}_{\cdot,j}\big)},
\]
where each column $\mathbf{X}_{\cdot,j}$ is the time trajectory of variable $j$. Thus, $L$ is the square root of the mean variance across variables. We add noise drawn from a determined distribution, such as Gaussian, to obtain corrupted observations. The corruption strength is controlled by a user-set \emph{relative noise level} $\sigma^2$, defined by
\[
\sigma^2 \;=\; \frac{\zeta^2}{L^2}.
\]
This normalisation adapts the absolute noise variance to the scale of the system, ensuring comparable noise levels across all problems.

%Then we construct a SIREN neural network and perform RKTV-INR on the noisy data. In our experiment, the network has three hidden layers, each with 80 units. During training, we used the Adam algorithm with a learning rate of $5 \times 10^{-4}$, and performed 3000 iterations. The selection of other hyperparameters varies by case. We use this network to estimate the noise-free state and first-order derivative. Eventually, these estimations are fed to SINDy to identify the dynamics.

%We also selected four classic methods introduced in \autoref{sect3} for comparison, including standard INR training, S-G Filter, TVR and Spline.

We construct a SIREN INR and train it on the noisy observations using the proposed RKTV-INR. In our experiments, the SIREN has three hidden layers with 80 neurons each. Optimisation uses Adam with a learning rate of $5\times 10^{-4}$ for 3000 iterations; other hyperparameters are tuned per case. The trained model provides denoised state estimates and first-order time derivatives (via automatic differentiation), which are then supplied to SINDy to identify the governing dynamics.

For comparison, we evaluate four baselines from \autoref{sect3}: standard INR fitting (without Runge-Kutta or TV terms) \cite{sitzmann2020implicit}, the Savitzky--Golay (S--G) filter \cite{1964Smoothing}, total variation regularisation (TVR) \cite{vogel2002computational}, and smoothing splines \cite{1975Smoothing}.

%The overall diagram of the experiment is shown as Figure \ref{fig5.9}. We first generate the system's state and first-order derivative data over a given time interval. Then, we add Gaussian white noise with different levels to them. During the training process, we use different algorithms to estimate the state and the first-order derivative, and compare these estimations with the ground truth.

%\begin{figure}[h]
%    \centering
%    \includegraphics[width=1\linewidth]{figures/diagram15.pdf}
%    \caption{The overall diagram of the experiment. }
%    \label{fig5.9}
%\end{figure}

\subsection{Metric}

We use the relative error between the estimated data and the ground-truth to measure the accuracy of the proposed method. The estimation error of $\mathbf{X}$ and $\dot{\mathbf{X}}$ are denoted by $e_{\mathbf{X}}$ and $e_{\dot{\mathbf{X}}}$ respectively, whose definition are as follows \cite{kaheman2022automatic}: 

\begin{align}
    e_{\mathbf{X}} = \frac{  \left| \mathbf{X} - \mathbf{\chi} \right|_F^2}{\left| \mathbf{X}\right|_F^2}, \ \     e_{\dot{\mathbf{X}}} = \frac{\left| \dot{\mathbf{X}} - \dot{\mathbf{\chi}} \right|_F^2}{ \left| \dot{\mathbf{X}} \right|_F^2},
\end{align}

\noindent where $\left| \cdot \right|_F^2$ means the Fréchet distance and $\mathbf{\chi}$ and $\dot{\mathbf{\chi}}$ are our estimated state and derivative. 

The estimated state and derivative data are then fed into the SINDy model to identify the dynamic system, provided by the coefficient matrix of the function library $\Xi$. To measure the identification ability of the SINDy model with our estimation data, the following indicator is defined as \cite{kaheman2022automatic}:

\begin{align}
    e_{\Xi} = \frac{\left| \Xi - \hat{\Xi} \right|_F^2}{\left| \Xi\right|_F^2},
\end{align}
where $\hat{\Xi}$ is our estimation. 

%In \autoref{section5.4}, we first add noise from different distributions to several simple dynamical systems to test the denoising capability of Runge–Kutta training and compare it with other baselines. In \autoref{section5.5}, using the Lorenz 63 and R\"{o}ssler systems as examples, we test the denoising capability of Runge–Kutta training under a range of relative noise values, feed our estimates into SINDy, and compare the performance with other baselines.

In \autoref{section5.4}, we inject noise from various distributions into several canonical dynamical systems to evaluate the denoising capability of Runge–Kutta training and compare it against baseline methods. In \autoref{section5.5}, using the Lorenz–63 and R\"{o}ssler systems, we sweep relative noise levels, apply SINDy to the denoised states and derivatives, and benchmark identification performance against the same baselines.

\subsection{Experimental results}\label{section5.4}

\subsubsection{State and Derivative Estimation}
\label{section5.4.1}

We selected four dynamical systems for testing: linear oscillator, cubic oscillator, Van der Pol oscillator and SEIR system. The linear oscillator is governed by the equation 

\begin{equation}
\begin{aligned}
    \dot{x}_1 = -0.1 x_1 + 3x_2\\
    \dot{x}_2 = -3x_1 - 0.1x_2
\end{aligned},
\end{equation}
with initial condition $x_1(0) = -2, \ x_2(0) = 2$. 
%We simulate the state data of the system on the time interval $[0,10]$ with a sampling density $h = 0.1$. The other three systems are all well-known and defined by standard canonical forms. Their definitions and the information regarding simulating the state data of the systems are provided in Appendix \ref{A2}. The average length scales $L$ of the four systems are $[1.27,\ 2.15 \times 10^{-1},\ 1.51,\ 2.84\times 10^{-1}]$, respectively.To generate the dataset, we add noise following Gaussian distribution with relative noise level $\sigma^2 = 10^{-2}$ to the ground-truth.

State data on the time interval $t\in[0,10]$ is simulated with time step $h=0.1$.
The remaining three systems follow standard canonical forms; their definitions
and simulation procedures are given in Appendix~\ref{A2}. The characteristic
length scales are $L=[1.27,\,2.15\times 10^{-1},\,1.51,\,2.84\times 10^{-1}]$.
To construct the dataset, we corrupt the ground truth with additive i.i.d.
Gaussian noise at a relative variance of $\sigma^{2}=10^{-2}$.

%The experimental results obtained using RKTV-INR are shown as Figure \ref{fig:state and derivative}, in which, the black solid line represents the ground-truth state data, the red dashed line represents noise data, the yellow dashed line represents state estimation data, and the blue dashed line represents derivative estimation data. As is evident from the figure, noisy data significantly interferes with the original data. Both the estimated state and the derivative derived from RKTV-INR proposed exhibit a high degree of consistency with the original data. This fully demonstrates that the proposed method can effectively mitigate the impact of noise in noisy scenarios, thereby retrieving valid data and accurately estimating its derivatives.

\begin{figure}[ht]
    \centering
    \includegraphics[width=0.9\linewidth]{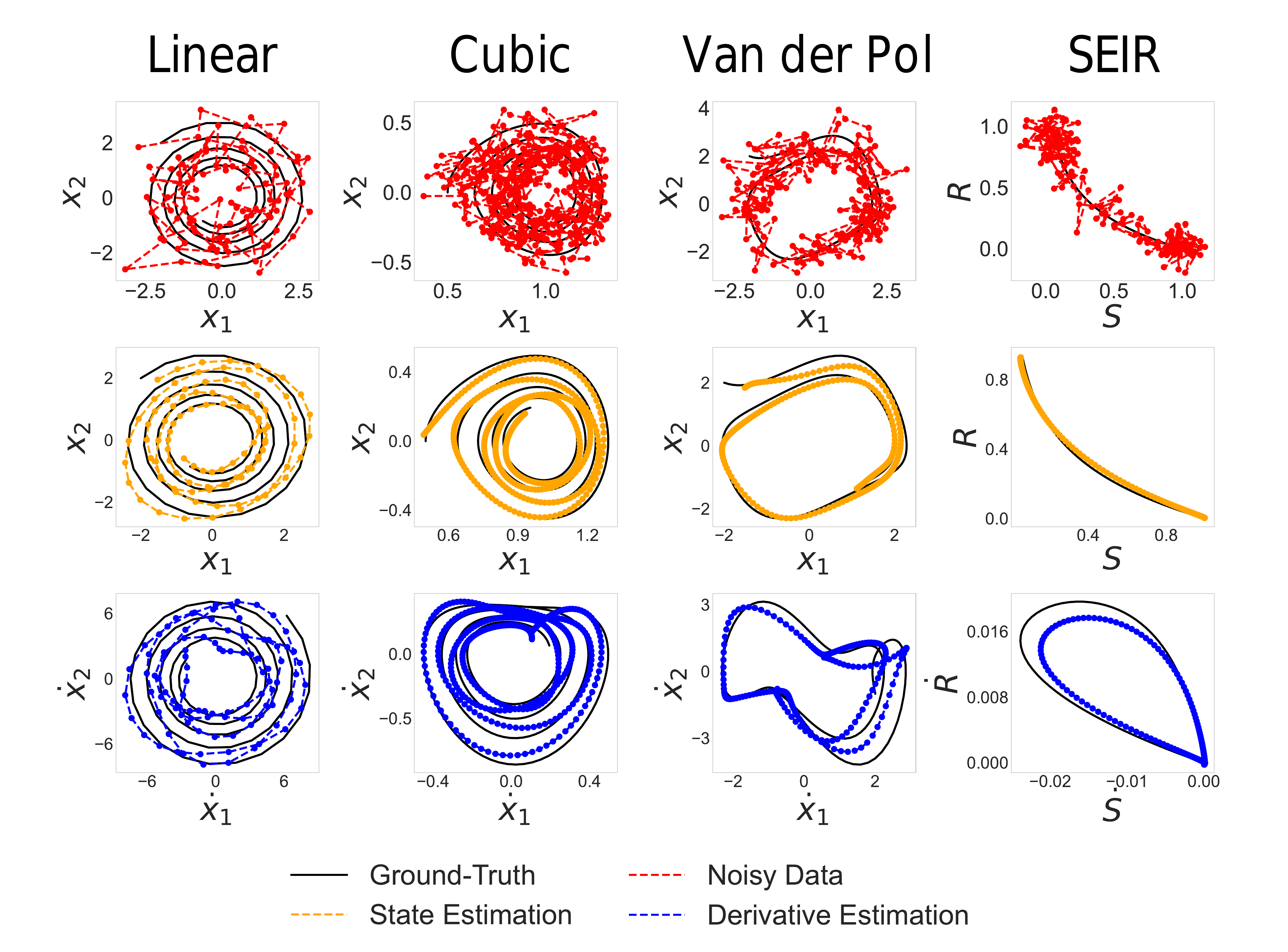}
    \caption{Estimation results given by RKTV-INR. Row-wise, the first row shows the generated noisy state data, while the second and third rows represent the state and derivative estimates obtained from RKTV-INR, respectively. Column-wise, the experiments are conducted on a linear oscillator, cubic oscillators, the Van der Pol oscillator, and the SEIR system.}
    \label{fig:state and derivative}
\end{figure}

Figure~\ref{fig:state and derivative} reports results obtained with RKTV-INR. 
The black solid curve denotes the ground-truth state, the red dashed curve the noisy observations, 
the yellow dashed curve the estimated state, and the blue dashed curve the estimated derivative. 
As shown, noise substantially distorts the observations; nonetheless, the RKTV-INR estimates for both 
state and derivative closely track the ground truth. This indicates that the proposed method effectively 
suppresses noise, recovers the underlying signal, and yields accurate derivative estimates.

%In addition, we have also compared RKTV-INR with the four baseline methods based on the metric $e_{\mathbf{X}}$ and $e_{\dot{\mathbf{X}}}$, and the experimental results are listed in Table \ref{table1}. It can be seen that the error of our method is the smallest for both state and derivative estimation. For state estimation, our method has an error that is on average $39.4 \%$ less than the results given by the S-G filter, the second-best method. For derivative estimation, compared to INR, the second-best approach, the error of our approach is $71.1 \%$ less. The result fully demonstrates that the method proposed in this article can effectively restore real state data and estimate its derivatives. 

In addition, we compare RKTV-INR with four baseline methods using the metrics $e_{\mathbf{X}}$ and $e_{\dot{\mathbf{X}}}$; the results are reported in Table~\ref{table1}. Our method achieves the lowest error for both state and derivative estimation. For state estimation, its error is on average $39.3\%$ lower than that of the S-G filter (the second-best method). For derivative estimation, it reduces error by $71.2\%$ relative to standard INR (the second-best). These results demonstrate that the proposed method effectively recovers the true state and accurately estimates its derivatives.

\begin{table}[ht]
\small
\centering
\caption{Comparison results of estimation errors of state and first-order derivatives between RKTV-INR and other baseline methods using metrics $e_{\mathbf{X}}$ and $e_{\dot{\mathbf{X}}}$. For state and derivative, our approach has errors that are $39.3 \%$ and $71.2 \%$ less than the second-best method, respectively.}
\label{table1}
\begin{tabular}{c|ccccc} % 6 列，全部居中对齐
\toprule
$e_{\mathbf{X}}$ & Standard INR &  RKTV-INR & S-G Filter & TVR & Spline\\
\hline
\hline  % 双线
Linear Oscillator & 3.35$e$-02 & \pmb{1.82$e$-02} & 2.29$e$-02 & 8.02$e$-02 & 3.68$e$-02 \\
Cubic Oscillator & 8.25$e$-03 & \pmb{1.06$e$-03} & 1.57$e$-03 & 2.63$e$-03 & 5.65$e$-03 \\
Van der Pol & 9.08$e$-03 & \pmb{7.93$e$-03} & 1.29$e$-02 & 2.02$e$-02 & 9.90$e$-03 \\
SEIR & 5.54$e$-03 & \pmb{1.04$e$-03} & 9.91$e$-03 & 7.67$e$-03 & 3.75$e$-03 \\
\hline
\hline  % 双线
$e_{\dot{\mathbf{X}}}$ & Standard INR & RKTV-INR & S-G Filter & TVR & Spline\\
\midrule
Linear Oscillator & 1.09$e$-01 & \pmb{4.00$e$-02} & 5.85$e$-02 & 1.61$e$-01 & 1.13$e$-01 \\
Cubic Oscillator & 1.53$e$-01 & \pmb{1.56$e$-02} & 2.92$e$-01 & 1.91$e$-01 & 1.69$e$-01 \\
Van der Pol & 1.05$e$-01 & \pmb{5.06$e$-02} & 1.29$e$-01 & 2.91$e$-01 & 1.07$e$-01 \\
SEIR & 1.06$e$-01 & \pmb{1.81$e$-02} & 4.77$e$-01 & 2.48$e$-01 & 3.42$e$-01 \\
\hline
\hline  % 双线
\end{tabular}
\end{table}

\subsubsection{Noise Distribution}

In the previous section, we tested the performance of RKTV-INR under the default that noise obeys a Gaussian distribution. However, noise may also be subordinate to other distributions, such as the uniform distribution \cite{clason2012fitting} and the Laplace distribution \cite{marks2007detection}.

\begin{figure}[ht]
    \centering
    \includegraphics[width=1\linewidth]{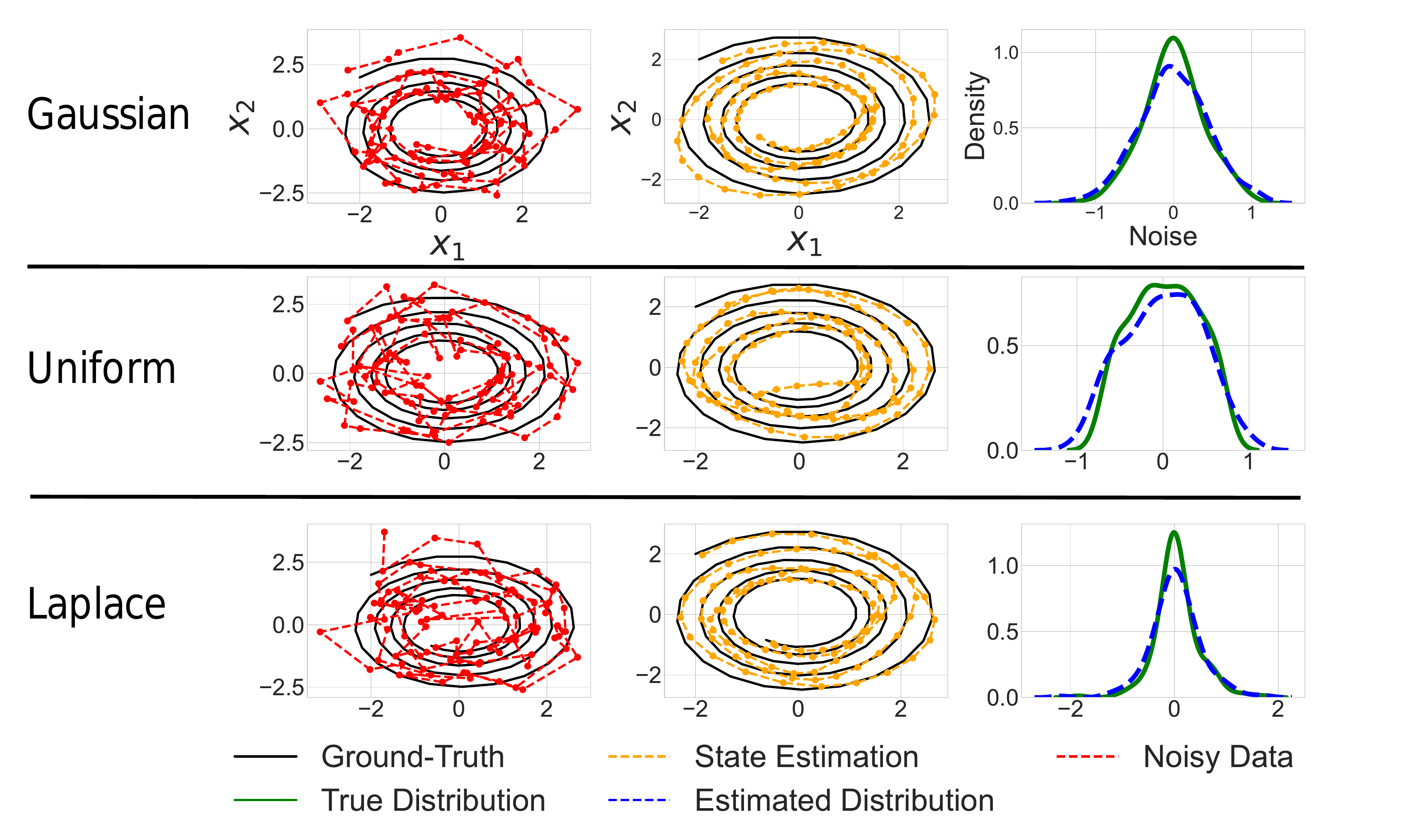}
    \caption{The identification results of noise with different distributions using RKTV-INR. Column-wise, the first column represents the noisy data, and the second and third columns represent the state and derivative estimation, respectively. Row-wise, we test the performance with noise following Gaussian, uniform and Laplace distributions, respectively.}
    \label{fig6}
\end{figure}

In this section, we use the linear oscillator as the experimental subject and add noise to the data with a relative noise level of $10^{-2}$, following a Gaussian distribution, uniform distribution, and Laplace distribution. Then, we denoise the three datasets separately using RKTV-INR and compare the noise distribution identified by the algorithm with the real noise distribution, as shown in Figure \ref{fig6}.

%As can be seen from Figure \ref{fig6},  for the three different distributions, the proposed method can well obtain the real state data, and the noise estimation can also fit the original distribution well. This indicates that the proposed method is capable of identifying noises in different forms.

Figure~\ref{fig6} shows that, across three distinct noise distributions, the proposed method accurately reconstructs the true state, while the estimated noise closely matches the generating distribution. This demonstrates robust noise identification across diverse noise types.

\subsubsection{Chaotic System Experiments}
\label{section5.5}

In this section, we select the Lorenz 63 System and R\"{o}ssler System as representative chaotic systems to evaluate the capability of RKTV-INR in handling complex data. For each system, we test the performance of RKTV-INR in denoising states and derivatives under different relative noise levels of Gaussian noise. The denoised estimates are then utilized in SINDy for dynamical system identification, and the results are compared with those obtained from other baselines.

\subsubsection*{Lorenz 63 System}

%The Lorenz 63 system is a well-known chaotic system characterised by sensitive dependence. Its governing equation and initial condition are attached to Appendix \ref{A2}. To improve the learning efficiency of our neural network, we first scale the state data of the system by 0.1. The average length scale $L$ of the scaled system is $8.34 \times 10^{-1}$. We simulate the system on the time interval $[0,10]$ with time step $h = 0.05$. Then, we add noise following the Gaussian distribution with different relative noise levels to test the RKTV-INR's performance under different circumstances. The relative noise levels include $\sigma^2$ = 1, and four lower scales $[10^{-1}$, $10^{-2}$, $10^{-3}$ and $10^{-4}]$, each combined with multipliers {1, 2/3, 1/3}, resulting in a total of 13 values.

The Lorenz–63 system is a canonical chaotic system exhibiting sensitive dependence on initial conditions; its governing equations and initial condition are provided in Appendix~\ref{A2}. To improve learning efficiency, we rescale the state by $0.1$, yielding an average length scale $L=8.34\times 10^{-1}$. We simulate over $t\in[0,10]$ with time step $h=0.05$. To assess robustness, we corrupt the trajectories with additive i.i.d. Gaussian noise at 13 relative variance levels:
\[
\sigma^2 \in \{1\} \cup \{\, m\times 10^{-k} : m\in\{1,\,2/3,\,1/3\},\ k\in\{1,2,3,4\} \,\}.
\]
These experiments evaluate the performance of Runge–Kutta training under varying noise intensities.

\begin{figure}[t]
    \centering
    \includegraphics[width=1\linewidth]{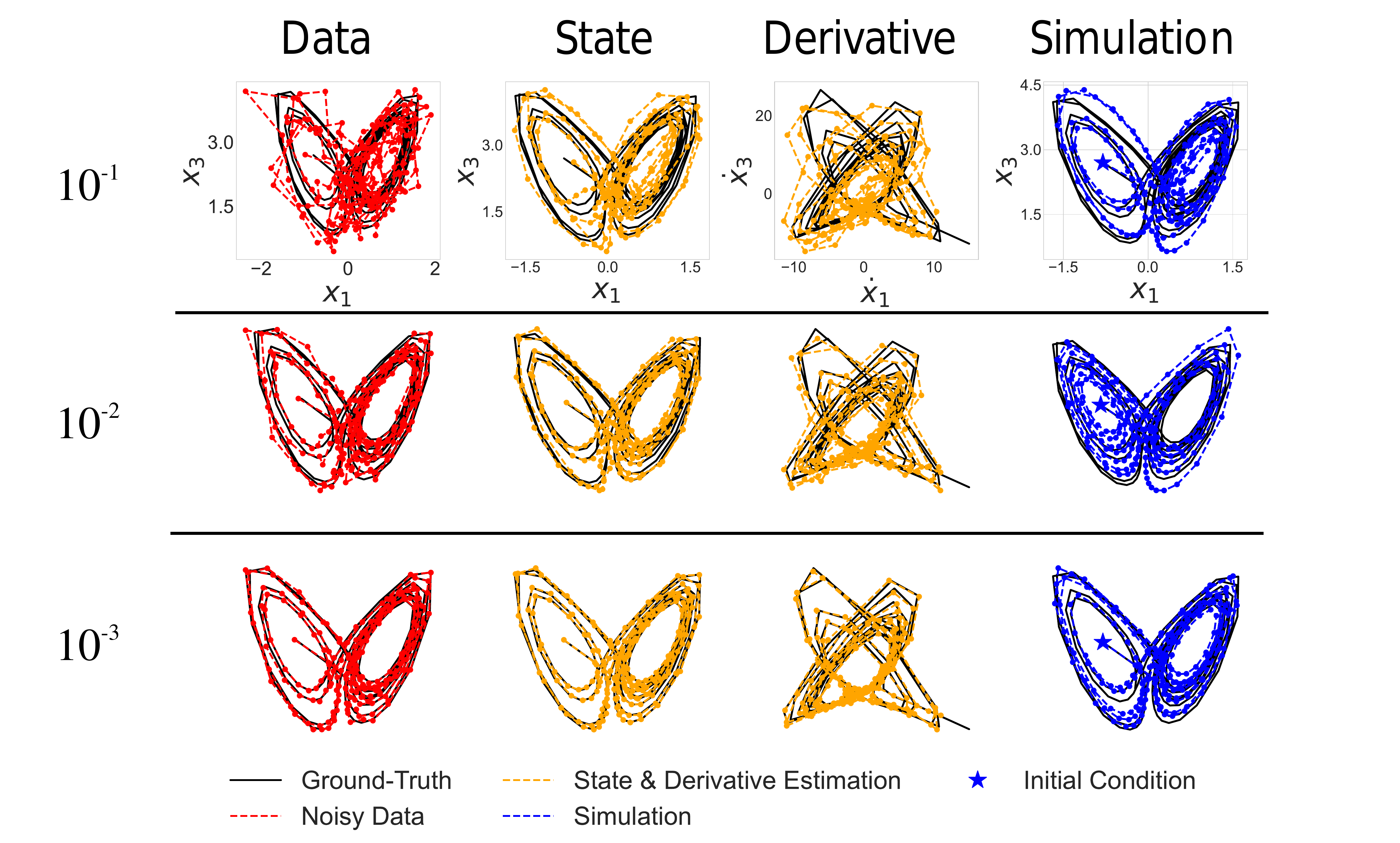}
    \caption{Experimental results of Lorenz 63 System using RKTV-INR under relative noise level $\sigma = 10^{-2},10^{-3},10^{-4}$. Column-wise, the first one is the noisy data, with the second and third representing the estimation of the state and derivative, respectively. The last column shows the simulated trajectory of the dynamic system provided by SINDy.}
    \label{fig:enter-label1}
\end{figure}

%The experimental results using RKTV-INR under relative noise levels $[10^{-1},10^{-2},10^{-3}]$ are shown as Figure \ref{fig:enter-label1}, in which, the first column shows data with different levels of noise, the second column is identified state data, the third column displays estimated derivative data, and the last column presents the simulated trajectory of the identified system given by SINDy. It can be seen that the method proposed in this paper can effectively obtain the ground-truth data and estimate the derivative for noise data of different levels. Finally, the compactness between the identified dynamical system and the real system is also very high, indicating that SINDy can accurately identify the governing equation from the estimation given by RKTV-INR.

Figure~\ref{fig:enter-label1} presents results for RKTV-INR at relative noise levels $[10^{-1},10^{-2},10^{-3}]$. 
The first column shows the noisy observations, the second the recovered state, the third the estimated derivatives, and the fourth the simulated trajectories of the system identified by SINDy. 
Across all noise levels, the proposed method recovers the underlying state and its derivatives with high accuracy. 
Moreover, the trajectories generated from the SINDy-identified model closely track the true dynamics, indicating that SINDy accurately infers the governing equations from the estimates produced by RKTV-INR.

\begin{figure}[ht]
    \centering
    \includegraphics[width=1\linewidth]{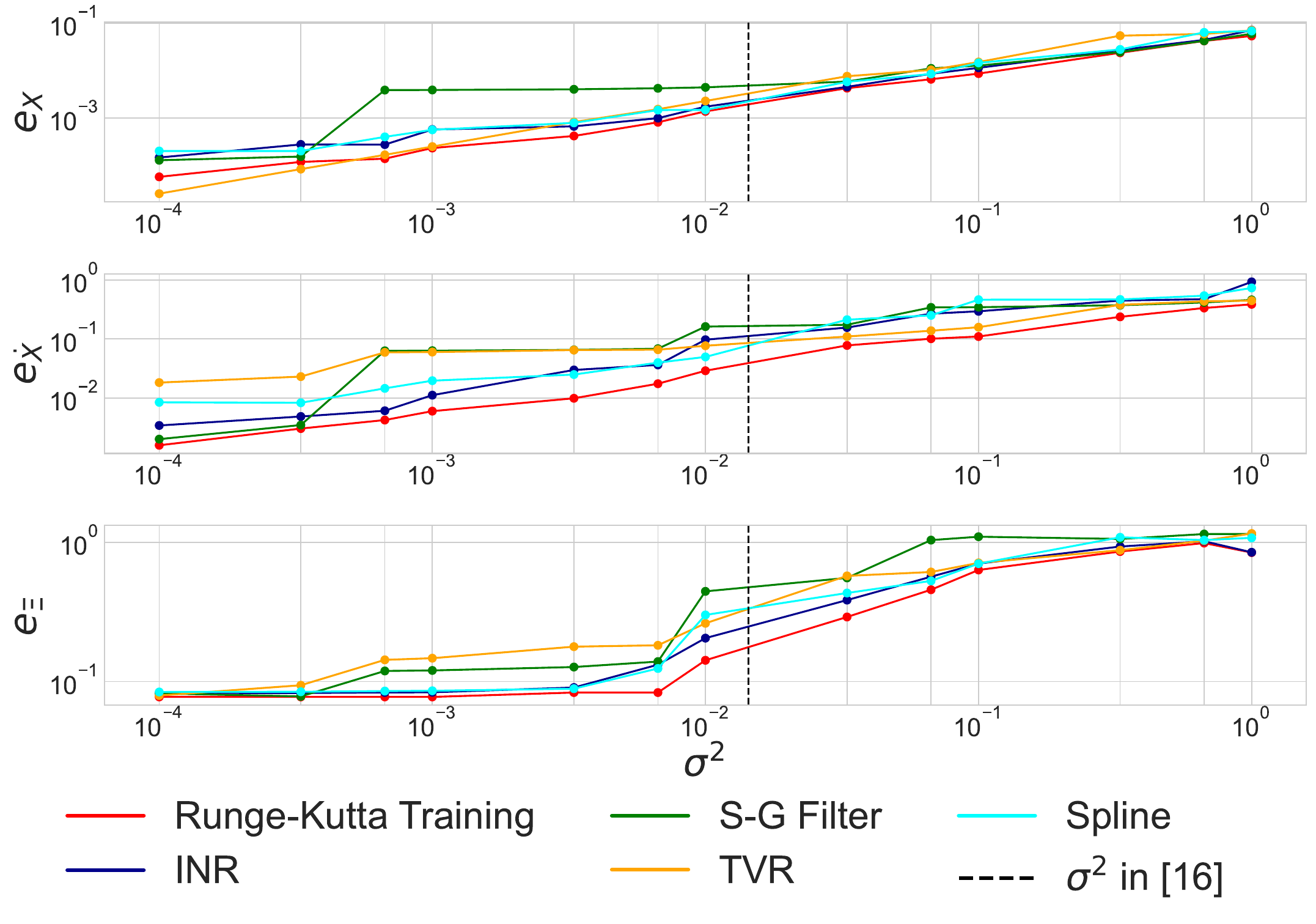}
    \caption{Comparison results of different methods on Lorenz 63 System. The relative noise levels include $\sigma^2$ = 1, and four lower scales $[10^{-1}$, $10^{-2}$, $10^{-3}$ and $10^{-4}]$, each combined with multipliers {1, 2/3, 1/3}. The black dashed vertical line represents the relative noise level of noise that is added in the paper where SINDy was published.}
    \label{fig:enter-label2}
\end{figure}

%We also compare RKTV-INR's performance to other baseline methods using metrics $e_{\mathbf{X}}$, $e_{\dot{\mathbf{X}}}$ and $e_{\Xi}$. As can be seen from Figure \ref{fig:enter-label2}, for different levels of relative noise, the error of the state data $e_{\mathbf{X}}$, the error of derivative estimation $e_{\dot{\mathbf{X}}}$, and the error of the identified coefficient matrix of the governing equation $e_{\Xi}$, are almost always the smallest. This result is especially noticeable when the relative noise level is between $10^{-3}$ and $10^{-1}$, where $e_{\mathbf{X}}$ and $e_{\dot{\mathbf{X}}}$ of our approach are on average $27.1\%$ and $58.6\%$ less than the second-best approach, INR. After feeding the estimation to SINDy, the identification error $e_{\Xi}$ of our approach is, on average, $19.5 \%$ less than that of INR. For situations with high or low noise intensity, although the advantages of our method are not so obvious, in most cases, the results are better than those of the comparative methods.

We compare RKTV-INR with baseline methods using the error metrics $e_{\mathbf{X}}$, $e_{\dot{\mathbf{X}}}$, and $e_{\Xi}$. As shown in Figure~\ref{fig:enter-label2}, across varying relative noise levels our method consistently achieves the lowest errors. In the regime between $10^{-3}$ and $10^{-1}$, $e_{\mathbf{X}}$ and $e_{\dot{\mathbf{X}}}$ are on average $27.1\%$ and $58.6\%$ lower than standard INR (the second-best method). After feeding the estimates into SINDy, the identification error $e_{\Xi}$ is on average $19.5\%$ lower than standard INR. At very high or very low noise intensities the margin narrows, but our approach remains competitive and, in most cases, superior to the alternatives.

%In the paper where SINDy was originally published \cite{brunton2016discovering}, they tested SINDy's performance under Gaussian noise with relative noise level $\sigma^2 = 1.4 \times 10^{-2}$. The black vertical dashed line in Figure \ref{fig:enter-label2} shows the relative noise level that SINDy proposed. As a data preprocessing step, they first use total variation regularisation to denoise the data before feeding it to SINDy. At the nearest relative noise level $\sigma^2 = 10^{-2}$, we find that our method’s denoising capability improves by $39.8 \%$ and $62.2 \%$ over total variation regularisation, corresponding to the state and the derivative, respectively. Compared with using total variation regularisation as the denoising method, our approach achieves a $45.7 \%$ higher accuracy in dynamic system identification after denoising. This fully demonstrates the improvements brought by our method over the previous SINDy framework.

In the original SINDy paper \cite{brunton2016discovering}, performance was evaluated under additive Gaussian noise with relative noise level $\sigma^2=1.4\times10^{-2}$, which is indicated by the black vertical dashed line in Figure~\ref{fig:enter-label2}. Their pipeline denoises the data using total variation regularisation before applying SINDy. At the relative noise level $\sigma^2=10^{-2}$ (nearest to $1.4\times10^{-2}$), our method yields lower errors in state and derivative estimation, $39.8\%$ and $62.2\%$, respectively, compared with total variation regularisation (second best method). When these estimates are passed to SINDy, the identification accuracy improves by $45.7\%$ relative to total variation regularisation. These results demonstrate a clear improvement over the standard SINDy workflow.

\subsubsection*{R\"{o}ssler System}

%The R\"{o}ssler system is another representative chaotic system. The model we simulate is also attached to Appendix \ref{A2}. The average length scale of the system is $L = 4.55$. We simulate the system over the time interval $[0,20]$ with a time step of $h = 0.05$. The choice of relative noise levels is the same as the one in the Lorenz 63 system, and we wish to test the performance of RKTV-INR under different relative noise levels.

The R\"ossler system is a canonical chaotic benchmark. The specific model used is provided in Appendix~\ref{A2}. The average length scale is $L=4.55$. We simulate trajectories over $t\in[0,20]$ with time step $h=0.05$. The relative noise levels match those used for the Lorenz-63 experiments, allowing us to assess the performance of RKTV-INR across varying noise intensities.

\begin{figure}[ht]
    \centering
    \includegraphics[width=1\linewidth]{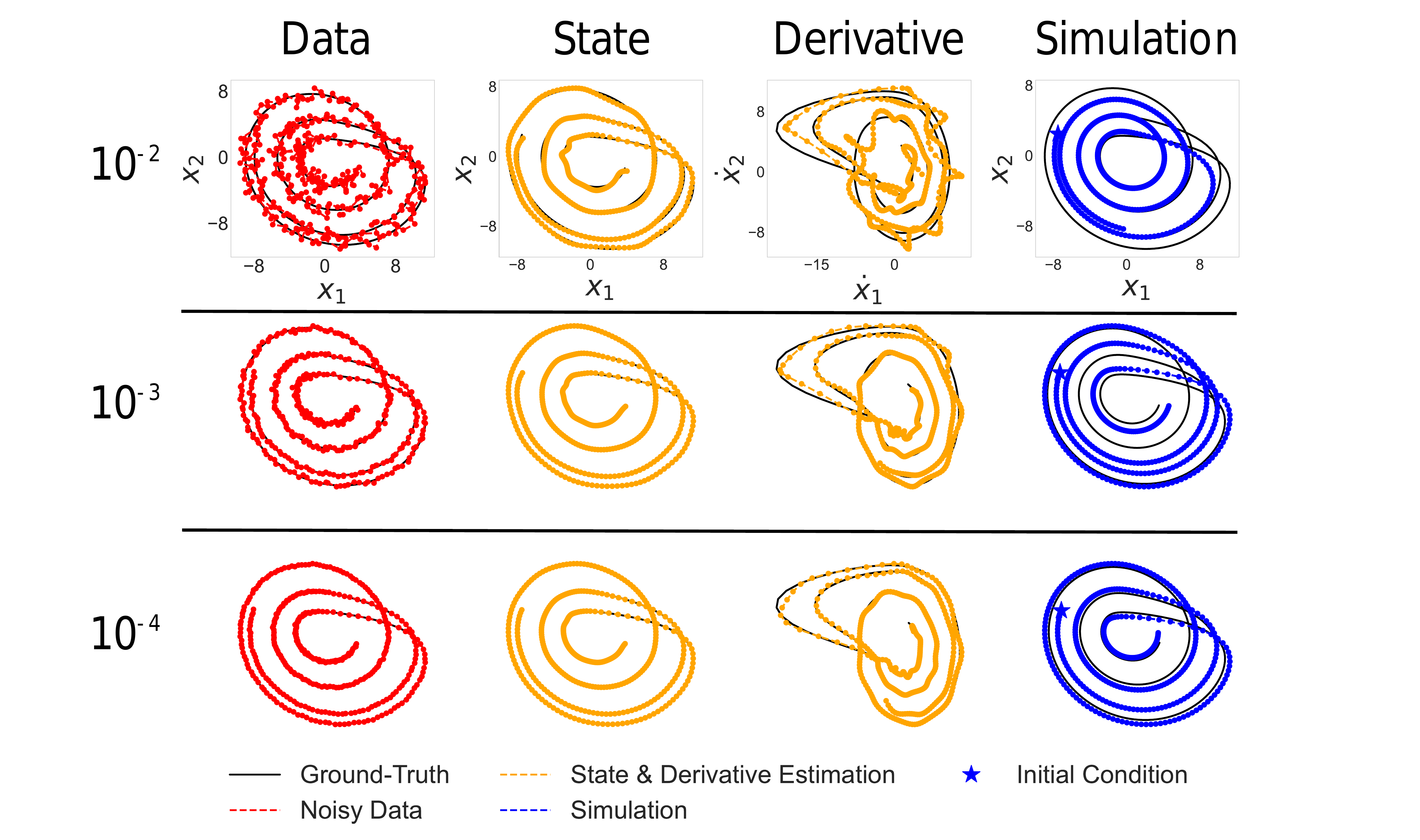}
    \caption{Experimental results of R\"{o}ssler System using RKTV-INR under relative noise levels $[10^{-2}, 10^{-3}, 10^{-4}]$. Column-wise, the first one is the noisy data, with the second and third representing the estimation of the state and derivative, respectively. The last column shows the simulated trajectory of the dynamic system provided by SINDy.}
    \label{fig:enter-label3}
\end{figure}

As illustrated in Figure \ref{fig:enter-label3}, RKTV-INR is still effective at accurately retrieving the ground-truth data and estimating derivatives, even for noisy data with varying intensity levels. Moreover, the identified dynamical system exhibits a high degree of consistency with the real system, while also satisfying the initial conditions.

\begin{figure}[ht]
    \centering
    \includegraphics[width=1\linewidth]{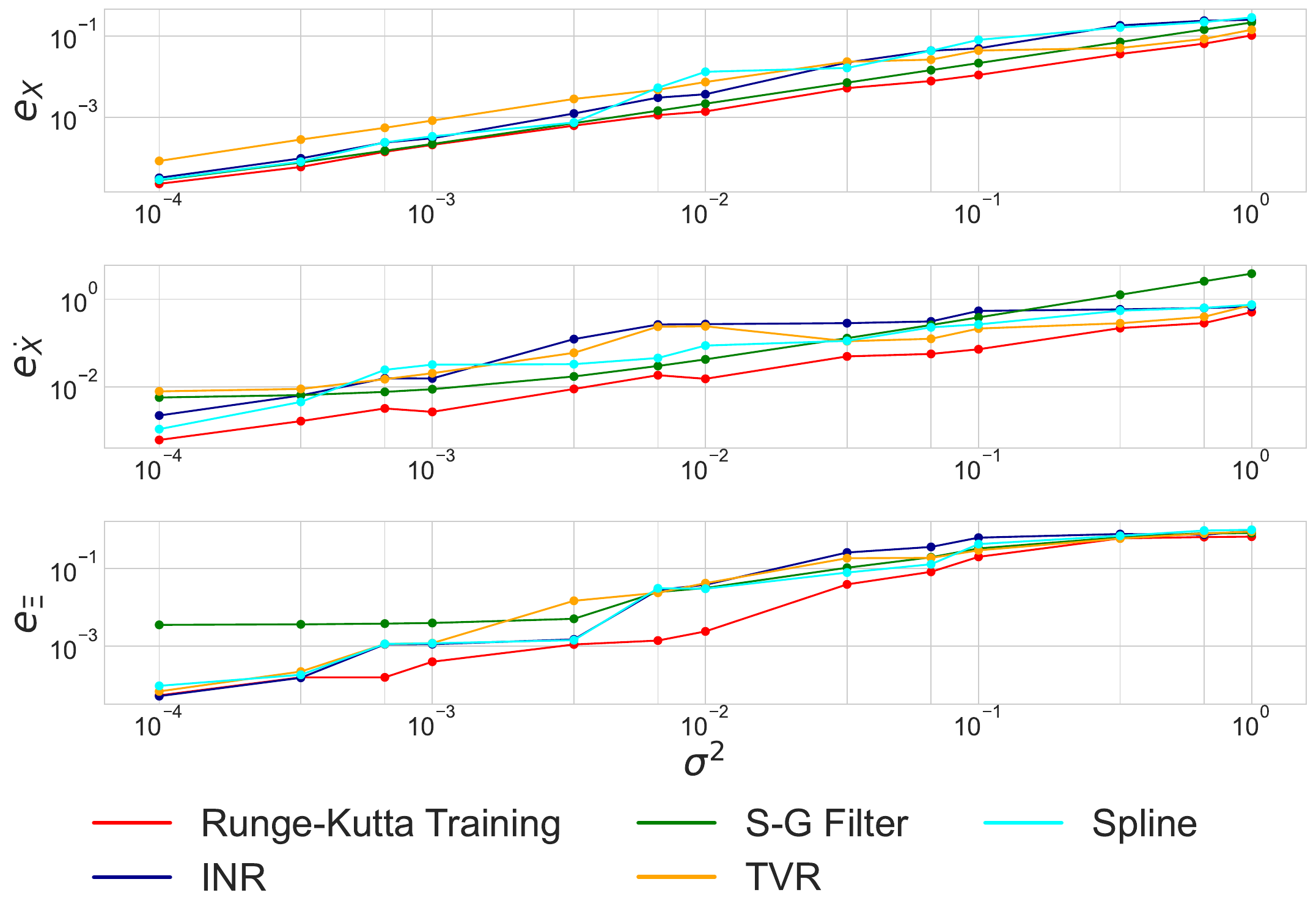}
    \caption{Comparison results of different methods on R\"{o}ssler System. The relative noise levels include $\sigma^2$ = 1, and four lower scales $[10^{-1}$, $10^{-2}$, $10^{-3}$ and $10^{-4}]$, each combined with multipliers {1, 2/3, 1/3}.}
    \label{fig:enter-label4}
\end{figure}

%As shown in Figure \ref{fig:enter-label4} where we compare our methods to baselines, across different relative noise levels, the proposed method achieves nearly the smallest errors in three key metrics: the state data error $e_{\mathbf{X}}$, the derivative estimation error $e_{\dot{\mathbf{X}}}$, and the error of the control equation's coefficient matrix $e_{\Xi}$. For state and derivative estimation, $e_{\mathbf{X}}$ and $e_{\dot{\mathbf{X}}}$ are $30.8 \%$ and $361 \%$ less than the second-best approach, S-G filter. For system identification, we can see that the advantages of our method become more apparent when the noise is relatively moderate, i.e. between $10^{-3}$ and $10^{-1}$. In this range, the identification error $e_{\Xi}$ is reduced by $72.8 \%$, $73.5 \%$, $73.6 \%$, and $59.4 \%$ when compared with INR, S-G filter, total variation regularisation and smoothing spline, respectively. This experiment fully demonstrates the advancement provided by our method in both data denoising and system identification.

Figure~\ref{fig:enter-label4} compares the proposed method with baselines across a range of relative noise levels. It attains the lowest—or near-lowest—errors on all three metrics: state error $e_{\mathbf{X}}$, derivative error $e_{\dot{\mathbf{X}}}$, and the governing-equation coefficient error $e_{\Xi}$. For state and derivative estimation, $e_{\mathbf{X}}$ and $e_{\dot{\mathbf{X}}}$ are $30.8\%$ and $70.9\%$ lower, respectively, than the second-best method (S-G filter). For system identification, the advantage is most pronounced at moderate noise levels $[10^{-3}$, –$10^{-1}]$, where $e_{\Xi}$ is reduced by $72.8\%$, $73.5\%$, $73.6\%$, and $59.4\%$ relative to standard INR, S-G filter, total variation regularisation, and smoothing spline, respectively. Overall, these results highlight the method’s strengths in both denoising and accurate system identification across diverse noise conditions.

\section{Summary}
\label{sect6}

%Data-driven approaches have been widely applied to the modeling of real-world dynamical systems. These models are often developed under the assumption of perfect data. However, in real world, the data collected from real systems is usually contaminated with noise, which can significantly reduce the accuracy of data-driven methods. For example, in the SINDy model, noise in the state data not only distorts the function library but also amplifies the error in first-order derivative estimation.

%To address this issue, we propose RKTV-INR, a novel two-step approach. It first accurately estimates the state and first-order derivatives from noisy observations, and then feeds the results into data-driven frameworks, such as SINDy. Our innovation lies in leveraging the recent implicit neural representation framework, specifically SIREN, to represent data as continuous functions. Furthermore, we incorporate Runge-Kutta integration and total variation as constraints to enhance both the accuracy and smoothness of the estimated results.

%Finally, through a series of experiments, we demonstrate that our method performs robustly across various dynamical system models, different noise distributions, and a wide range of relative noise levels. We also compare our approach with several baseline methods and validate its effectiveness.

Data-driven methods are widely used to model real-world dynamical systems, but they are often developed under an implicit assumption of noise-free data. In practice, measurements are contaminated by noise, which can markedly degrade performance. For example, in SINDy, noisy state measurements corrupt the candidate function library and exacerbate errors in derivative estimation.

To address this, we introduce RKTV-INR, a two-step procedure. First, from noisy observations we recover accurate state trajectories and their first-order derivatives by fitting an implicit neural representation (INR) that treats the data as a continuous function, while enforcing Runge-Kutta integration residuals and total variation regularisation to promote consistency and smoothness. Second, we feed these recovered states and derivatives into downstream data-driven frameworks, such as SINDy. Extensive experiments across multiple dynamical systems, noise distributions, and a broad range of relative noise levels show that the proposed approach is robust and consistently outperforms strong baselines in both denoising and system identification.

In future work, we may consider using  RKTV-INR as a data preprocessing algorithm and combining it with other data-driven dynamical system methods. Finally, we also recognize that many real-world measurements are recorded on irregular time grids. Therefore, RK-Training can be further extended to handle data collected on non-uniform time domains.

\section{Acknowledgement}

This work was carried out by Jiaqi Yao with Hemanth Saratchandran, John Maclean and Lewis Mitchell as his supervisors while pursuing an Honours degree at the University of Adelaide. He received a research scholarship from the Australian Institute for Machine Learning to support the completion of this work.

\bibliographystyle{IEEEtran}
\bibliography{ref.bib}

\begin{thebibliography}{10}
\providecommand{\url}[1]{#1}
\csname url@samestyle\endcsname
\providecommand{\newblock}{\relax}
\providecommand{\bibinfo}[2]{#2}
\providecommand{\BIBentrySTDinterwordspacing}{\spaceskip=0pt\relax}
\providecommand{\BIBentryALTinterwordstretchfactor}{4}
\providecommand{\BIBentryALTinterwordspacing}{\spaceskip=\fontdimen2\font plus
\BIBentryALTinterwordstretchfactor\fontdimen3\font minus \fontdimen4\font\relax}
\providecommand{\BIBforeignlanguage}[2]{{%
\expandafter\ifx\csname l@#1\endcsname\relax
\typeout{** WARNING: IEEEtran.bst: No hyphenation pattern has been}%
\typeout{** loaded for the language `#1'. Using the pattern for}%
\typeout{** the default language instead.}%
\else
\language=\csname l@#1\endcsname
\fi
#2}}
\providecommand{\BIBdecl}{\relax}
\BIBdecl

\bibitem{ghadami2022data}
A.~Ghadami and B.~I. Epureanu, ``Data-driven prediction in dynamical systems: recent developments,'' \emph{Philosophical Transactions of the Royal Society A}, vol. 380, no. 2229, p. 20210213, 2022.

\bibitem{north2023review}
J.~S. North, C.~K. Wikle, and E.~M. Schliep, ``A review of data-driven discovery for dynamic systems,'' \emph{International Statistical Review}, vol.~91, no.~3, pp. 464--492, 2023.

\bibitem{chen2021physics}
Z.~Chen, Y.~Liu, and H.~Sun, ``Physics-informed learning of governing equations from scarce data,'' \emph{Nature communications}, vol.~12, no.~1, p. 6136, 2021.

\bibitem{zhang_subtsbr_2021}
S.~Zhang and G.~Lin, ``{SubTSBR} to tackle high noise and outliers for data-driven discovery of differential equations,'' \emph{Journal of Computational Physics}, vol. 428, p. 109962, 2021.

\bibitem{he2020seir}
S.~He, Y.~Peng, and K.~Sun, ``{SEIR modeling of the COVID-19 and its dynamics},'' \emph{Nonlinear dynamics}, vol. 101, pp. 1667--1680, 2020.

\bibitem{tu2013dynamic}
J.~H. Tu, ``Dynamic mode decomposition: Theory and applications,'' Ph.D. dissertation, Princeton University, 2013.

\bibitem{raissi2019physics}
M.~Raissi, P.~Perdikaris, and G.~E. Karniadakis, ``Physics-informed neural networks: A deep learning framework for solving forward and inverse problems involving nonlinear partial differential equations,'' \emph{Journal of Computational physics}, vol. 378, pp. 686--707, 2019.

\bibitem{2023Random}
Y.~Liu, S.~G. Mccalla, and H.~Schaeffer, ``Random feature models for learning interacting dynamical systems,'' \emph{Proceedings of the Royal Society A: Mathematical, Physical $\&$ Engineering Sciences.}, vol. 479, no. 2275, pp. 1--23, 2023.

\bibitem{yamada2024spatial}
H.~Yamada, ``Spatial smoothing using graph laplacian penalized filter,'' \emph{Spatial Statistics}, vol.~60, p. 100799, 2024.

\bibitem{brunton2016discovering}
S.~L. Brunton, J.~L. Proctor, and J.~N. Kutz, ``Discovering governing equations from data by sparse identification of nonlinear dynamical systems,'' \emph{Proceedings of the national academy of sciences}, vol. 113, no.~15, pp. 3932--3937, 2016.

\bibitem{viknesh2024adam}
S.~Viknesh, Y.~Tatari, and A.~Arzani, ``{ADAM-SINDy}: An efficient optimization framework for parameterized nonlinear dynamical system identification,'' \emph{arXiv preprint arXiv:2410.16528}, 2024.

\bibitem{fukami2021sparse}
K.~Fukami, T.~Murata, K.~Zhang, and K.~Fukagata, ``Sparse identification of nonlinear dynamics with low-dimensionalized flow representations,'' \emph{Journal of Fluid Mechanics}, vol. 926, p. A10, 2021.

\bibitem{carderera2021cindy}
A.~Carderera, S.~Pokutta, C.~Sch{\"u}tte, and M.~Weiser, ``{CINDy}: Conditional gradient-based identification of non-linear {Dynamics--Noise-robust} recovery,'' \emph{arXiv preprint arXiv:2101.02630}, 2021.

\bibitem{garcia2015dealing}
S.~Garc{\'\i}a, J.~Luengo, F.~Herrera, S.~Garc{\'\i}a, J.~Luengo, and F.~Herrera, ``Dealing with noisy data,'' \emph{Data Preprocessing in Data Mining}, vol.~72, pp. 107--145, 2015.

\bibitem{wu2021challenges}
Z.~Wu, S.~L. Brunton, and S.~Revzen, ``Challenges in dynamic mode decomposition,'' \emph{Journal of the Royal Society Interface}, vol.~18, no. 185, p. 20210686, 2021.

\bibitem{dawson2016characterizing}
S.~T. Dawson, M.~S. Hemati, M.~O. Williams, and C.~W. Rowley, ``Characterizing and correcting for the effect of sensor noise in the dynamic mode decomposition,'' \emph{Experiments in Fluids}, vol.~57, pp. 1--19, 2016.

\bibitem{satyadharma2024assessing}
A.~Satyadharma, M.-J. Chern, H.-C. Kan, H.~Harinaldi, and J.~Julian, ``Assessing physics-informed neural network performance with sparse noisy velocity data,'' \emph{Physics of Fluids}, vol.~36, no.~10, p. 103619, 2024.

\bibitem{zou2025uncertainty}
Z.~Zou, X.~Meng, and G.~E. Karniadakis, ``{Uncertainty quantification for noisy inputs--outputs in physics-informed neural networks and neural operators},'' \emph{Computer Methods in Applied Mechanics and Engineering}, vol. 433, p. 117479, 2025.

\bibitem{wentz2023derivative}
J.~Wentz and A.~Doostan, ``{Derivative-based SINDy (DSINDy): Addressing the challenge of discovering governing equations from noisy data},'' \emph{Computer Methods in Applied Mechanics and Engineering}, vol. 413, p. 116096, 2023.

\bibitem{rudy2019deep}
S.~H. Rudy, J.~N. Kutz, and S.~L. Brunton, ``Deep learning of dynamics and signal-noise decomposition with time-stepping constraints,'' \emph{Journal of Computational Physics}, vol. 396, pp. 483--506, 2019.

\bibitem{hemati2017biasing}
M.~S. Hemati, C.~W. Rowley, E.~A. Deem, and L.~N. Cattafesta, ``{De-biasing the dynamic mode decomposition for applied Koopman spectral analysis of noisy datasets},'' \emph{Theoretical and Computational Fluid Dynamics}, vol.~31, pp. 349--368, 2017.

\bibitem{thanasutives2023noise}
P.~Thanasutives, T.~Morita, M.~Numao, and K.-i. Fukui, ``{Noise-aware physics-informed machine learning for robust PDE discovery},'' \emph{Machine Learning: Science and Technology}, vol.~4, no.~1, p. 015009, 2023.

\bibitem{wang2023simultaneous}
J.~Wang, J.~Moreira, Y.~Cao, and R.~B. Gopaluni, ``Simultaneous digital twin identification and signal-noise decomposition through modified generalized sparse identification of nonlinear dynamics,'' \emph{Computers $\&$ Chemical Engineering}, vol. 177, p. 108294, 2023.

\bibitem{kaheman2022automatic}
K.~Kaheman, S.~L. Brunton, and J.~N. Kutz, ``Automatic differentiation to simultaneously identify nonlinear dynamics and extract noise probability distributions from data,'' \emph{Machine Learning: Science and Technology}, vol.~3, no.~1, p. 015031, 2022.

\bibitem{goyal2022discovery}
P.~Goyal and P.~Benner, ``{Discovery of nonlinear dynamical systems using a Runge--Kutta inspired dictionary-based sparse regression approach},'' \emph{Proceedings of the Royal Society A}, vol. 478, no. 2262, p. 20210883, 2022.

\bibitem{forootani2023robust}
A.~Forootani, P.~Goyal, and P.~Benner, ``A robust sparse identification of nonlinear dynamics approach by combining neural networks and an integral form,'' \emph{Engineering Applications of Artificial Intelligence}, vol. 149, p. 110360, 2025.

\bibitem{zhai2023parameter}
W.~Zhai, D.~Tao, and Y.~Bao, ``{Parameter estimation and modeling of nonlinear dynamical systems based on Runge--Kutta physics-informed neural network},'' \emph{Nonlinear Dynamics}, vol. 111, no.~22, pp. 21\,117--21\,130, 2023.

\bibitem{zhai2024state}
W.~Zhai, Y.~Bao, and D.~Tao, ``{State space model-based Runge--Kutta gated recurrent unit networks for structural response prediction},'' \emph{Nonlinear Dynamics}, vol. 112, no.~24, pp. 21\,901--21\,921, 2024.

\bibitem{lubansky2006general}
A.~Lubansky, Y.~L. Yeow, Y.-K. Leong, S.~R. Wickramasinghe, and B.~Han, ``A general method of computing the derivative of experimental data,'' \emph{AICHE Journal}, vol.~52, no.~1, pp. 323--332, 2006.

\bibitem{kostelich1993noise}
E.~J. Kostelich and T.~Schreiber, ``Noise reduction in chaotic time-series data: A survey of common methods,'' \emph{Physical Review E}, vol.~48, no.~3, p. 1752, 1993.

\bibitem{kasac2018algebraic}
J.~Kasac, D.~Majetic, and D.~Brezak, ``An algebraic approach to on-line signal denoising and derivatives estimation,'' \emph{Journal of the Franklin Institute}, vol. 355, no.~15, pp. 7799--7825, 2018.

\bibitem{green_nonparametric_1993}
P.~J. Green and B.~W. Silverman, \emph{Nonparametric Regression and Generalized Linear Models: A Roughness Penalty Approach}.\hskip 1em plus 0.5em minus 0.4em\relax New York: Chapman and Hall/CRC, 1993.

\bibitem{1964Smoothing}
A.~Savitzky, ``Smoothing and differentiation of data by simplified least squares procedures.'' \emph{Analytical Chemistry}, vol.~36, no.~8, pp. 1627--2639, 1964.

\bibitem{schafer2011savitzky}
R.~W. Schafer, ``What is a {Savitzky-Golay} filter?[lecture notes],'' \emph{IEEE Signal Processing Magazine}, vol.~28, no.~4, pp. 111--117, 2011.

\bibitem{ahnert2007numerical}
K.~Ahnert and M.~Abel, ``Numerical differentiation of experimental data: local versus global methods,'' \emph{Computer Physics Communications}, vol. 177, no.~10, pp. 764--774, 2007.

\bibitem{stickel_data_2010}
J.~J. Stickel, ``Data smoothing and numerical differentiation by a regularization method,'' \emph{Computers $\&$ Chemical Engineering}, vol.~34, no.~4, pp. 467--475, 2010.

\bibitem{sun_physics-informed_2021}
F.~Sun, Y.~Liu, and H.~Sun, ``Physics-informed {Spline} {Learning} for {Nonlinear} {Dynamics} {Discovery},'' \emph{10.48550/arXiv.2105.02368}, 2021.

\bibitem{1975Smoothing}
P.~Craven and G.~Wahba, ``Smoothing noisy data with spline functions,'' \emph{Numerische Mathematick}, vol.~31, no.~5, pp. 377--403, 1979.

\bibitem{tikhonov1963regularization}
A.~N. Tikhonov, ``Regularization of incorrectly posed problems,'' \emph{Soviet Mathematics Doklady}, vol.~4, pp. 1624--1627, 1963.

\bibitem{vogel2002computational}
C.~R. Vogel, \emph{Computational methods for inverse problems}.\hskip 1em plus 0.5em minus 0.4em\relax SIAM, 2002.

\bibitem{chartrand2011numerical}
R.~Chartrand, ``Numerical differentiation of noisy, nonsmooth data,'' \emph{International Scholarly Research Notices}, vol. 2011, no.~1, p. 164564, 2011.

\bibitem{zhu2019seismic}
W.~Zhu, S.~M. Mousavi, and G.~C. Beroza, ``Seismic signal denoising and decomposition using deep neural networks,'' \emph{IEEE Transactions on Geoscience and Remote Sensing}, vol.~57, no.~11, pp. 9476--9488, 2019.

\bibitem{fan2020vibration}
G.~Fan, J.~Li, and H.~Hao, ``Vibration signal denoising for structural health monitoring by residual convolutional neural networks,'' \emph{Measurement}, vol. 157, p. 107651, 2020.

\bibitem{yu2019deep}
S.~Yu, J.~Ma, and W.~Wang, ``Deep learning for denoising,'' \emph{Geophysics}, vol.~84, no.~6, pp. V333--V350, 2019.

\bibitem{park2019deepsdf}
J.~J. Park, P.~Florence, J.~Straub, R.~Newcombe, and S.~Lovegrove, ``Deepsdf: Learning continuous signed distance functions for shape representation,'' in \emph{Proceedings of the IEEE/CVF Conference on Computer Vision and Pattern Recognition}, 16--20 June 2019, pp. 165--174.

\bibitem{mildenhall2021nerf}
B.~Mildenhall, P.~P. Srinivasan, M.~Tancik, J.~T. Barron, R.~Ramamoorthi, and R.~Ng, ``Nerf: {Representing} scenes as neural radiance fields for view synthesis,'' \emph{Communications of the ACM}, vol.~65, no.~1, pp. 99--106, 2021.

\bibitem{baydin2018automatic}
A.~G. Baydin, B.~A. Pearlmutter, A.~A. Radul, and J.~M. Siskind, ``Automatic differentiation in machine learning: a survey,'' \emph{Journal of machine learning research}, vol.~18, no. 153, pp. 1--43, 2018.

\bibitem{epperson2013introduction}
J.~F. Epperson, \emph{An introduction to numerical methods and analysis}.\hskip 1em plus 0.5em minus 0.4em\relax John Wiley $\&$ Sons, 2013.

\bibitem{kwapien2012physical}
J.~Kwapie{\'n} and S.~Dro{\.z}d{\.z}, ``Physical approach to complex systems,'' \emph{Physics Reports}, vol. 515, no. 3-4, pp. 115--226, 2012.

\bibitem{10.1111/j.2517-6161.1996.tb02080.x}
R.~Tibshirani, ``Regression shrinkage and selection via the {Lasso},'' \emph{Journal of the Royal Statistical Society: Series B (Methodological)}, vol.~58, no.~1, pp. 267--288, 2018.

\bibitem{Silverman1984A}
B.~W. Silverman, ``A fast and efficient cross-validation method for smoothing parameter choice in spline regression,'' \emph{Publications of the American Statistical Association}, vol.~79, no. 387, pp. 584--589, 1984.

\bibitem{2012Local}
Z.~Shang and G.~Cheng, ``Local and global asymptotic inference in smoothing spline models,'' \emph{The Annals of Statistics}, vol.~41, no.~5, pp. 2608–--2638, 2012.

\bibitem{bishop2023deep}
C.~M. Bishop and H.~Bishop, \emph{Deep learning: Foundations and concepts}.\hskip 1em plus 0.5em minus 0.4em\relax Springer Nature, 2023.

\bibitem{ramasinghe2023effectiveness}
S.~Ramasinghe, H.~Saratchandran, V.~Shevchenko, and S.~Lucey, ``On the effectiveness of neural priors in modeling dynamical systems,'' \emph{arXiv preprint arXiv:2303.05728}, 2023.

\bibitem{saratchandran2024sampling}
H.~Saratchandran, S.~Ramasinghe, V.~Shevchenko, A.~Long, and S.~Lucey, ``A sampling theory perspective on activations for implicit neural representations,'' in \emph{Proceedings of the 41st International Conference on Machine Learning}, vol. 235, 21--27 July 2024, pp. 43\,422--43\,444.

\bibitem{saratchandran2023curvature}
H.~Saratchandran, S.-F. Chng, S.~Ramasinghe, L.~MacDonald, and S.~Lucey, ``Curvature-aware training for coordinate networks,'' in \emph{Proceedings of the IEEE/CVF International Conference on Computer Vision}, 1-6 October 2023, pp. 13\,328--13\,338.

\bibitem{saratchandran2024activation}
H.~Saratchandran, S.~Ramasinghe, and S.~Lucey, ``From activation to initialization: Scaling insights for optimizing neural fields,'' in \emph{Proceedings of the IEEE/CVF Conference on Computer Vision and Pattern Recognition}, 16-22 June 2024, pp. 413--422.

\bibitem{sitzmann2020implicit}
V.~Sitzmann, J.~Martel, A.~Bergman, D.~Lindell, and G.~Wetzstein, ``Implicit neural representations with periodic activation functions,'' \emph{Advances in neural information processing systems}, vol.~33, pp. 7462--7473, 2020.

\bibitem{saragadam2023wire}
V.~Saragadam, D.~LeJeune, J.~Tan, G.~Balakrishnan, A.~Veeraraghavan, and R.~G. Baraniuk, ``Wire: Wavelet implicit neural representations,'' in \emph{Proceedings of the IEEE/CVF Conference on Computer Vision and Pattern Recognition}, 17--24 June 2023, pp. 18\,507--18\,516.

\bibitem{saitta2024implicit}
S.~Saitta, M.~Carioni, S.~Mukherjee, C.-B. Sch{\"o}nlieb, and A.~Redaelli, ``{Implicit neural representations for unsupervised super-resolution and denoising of 4D flow MRI},'' \emph{Computer methods and programs in biomedicine}, vol. 246, p. 108057, 2024.

\bibitem{kim2022zero}
C.~Kim, J.~Lee, and J.~Shin, ``Zero-shot blind image denoising via implicit neural representations,'' \emph{arXiv preprint arXiv:2204.02405}, 2022.

\bibitem{10.5555/3294996.3295182}
W.~M. Czarnecki, S.~Osindero, M.~Jaderberg, G.~Swirszcz, and R.~Pascanu, ``Sobolev training for neural networks,'' in \emph{Proceedings of the 31st International Conference on Neural Information Processing Systems}, ser. NIPS'17.\hskip 1em plus 0.5em minus 0.4em\relax Red Hook, NY, USA: Curran Associates Inc., 2017, p. 4281–4290.

\bibitem{clason2012fitting}
C.~Clason, ``$l_{\infty}$ fitting for inverse problems with uniform noise,'' \emph{Inverse Problems}, vol.~28, no.~10, p. 104007, 2012.

\bibitem{marks2007detection}
R.~J. Marks, G.~L. Wise, D.~G. Haldeman, and J.~L. Whited, ``Detection in laplace noise,'' \emph{IEEE Transactions on Aerospace and Electronic Systems}, no.~6, pp. 866--872, 2007.

\end{thebibliography}

\clearpage
\section{Supplementary material}

\subsection{Derivation of Total Variation}
\label{A1}

In \autoref{section4.2}, we mentioned that the smoothness loss function is identical to the total variation of the second-order derivative in the limit $h \rightarrow 0$. This appendix provides a mathematical analysis of this approximation. As a recap, the smoothness loss function is 

\begin{align}
\mathcal{L}_3 &=  \frac{1}{m-1} \sum_{i=1}^{m-1} \left| \frac{\mathrm{d}^2 \mathbf{\chi}_{\boldsymbol{\theta}}}{\mathrm{d} t^2}(t_{i+1}) - \frac{\mathrm{d}^2\mathbf{\chi}_{\boldsymbol{\theta}}}{\mathrm{d} t^2}(t_{i})\right|_2^2 \nonumber \\
&= \frac{h^2}{m-1} \sum_{i=1}^{m-1} \left| \frac{\frac{\mathrm{d}^2 \mathbf{\chi}_{\boldsymbol{\theta}}}{\mathrm{d} t^2}(t_{i+1}) - \frac{\mathrm{d}^2\mathbf{\chi}_{\boldsymbol{\theta}}}{\mathrm{d} t^2}(t_{i})}{h}\right|_2^2 \nonumber \\
&\mathrm{In \ the \ limit} \ h \rightarrow 0,\ \mathrm{by \ central \ difference} \nonumber \\
&= \frac{h^2}{m-1} \sum_{i=1}^{m-1} \left| \frac{\mathrm{d}^3 \mathbf{\chi}_{\boldsymbol{\theta}}}{\mathrm{d} t^3}(\frac{t_{i+1}+ t_{i}}{2}) \right|_2^2 \nonumber \\
&= \frac{h}{m-1} \sum_{i=1}^{m-1} \left| \frac{\mathrm{d}^3 \mathbf{\chi}_{\boldsymbol{\theta}}}{\mathrm{d} t^3}(\frac{t_{i+1}+ t_{i}}{2}) \right|_2^2 \cdot h \nonumber \\
&\mathrm{In \ the \ limit} \ h \rightarrow 0,\ \mathrm{by \ midpoint \ rule} \nonumber \\
&= \frac{h}{m-1} \int_a^b \left[ \frac{\mathrm{d}}{\mathrm{d} t} \left(\frac{\mathrm{d^2 \chi_{\boldsymbol{\theta}}}}{\mathrm{d} t^2} \right) 
\right]^2 \mathrm{d} t.
\end{align}

\subsection{Dynamic System Models}
\label{A2}
In \autoref{section5.4}, we demonstrate the experimental results of RKTV-INR on different dynamic systems. This appendix provides the mathematical models of each dynamical system, along with the initial conditions used in our simulations.

\subsubsection{Cubic Oscillator}

The cubic oscillator is governed by the equation 

\begin{equation}
\begin{aligned}
    \dot{x}_1 &= x_2\\
    \dot{x}_2 &= -\delta x_2 - \alpha x_1 - \beta x_1^3
\end{aligned},
\end{equation}
where $\delta = 0.1,\alpha = -1, \beta = 1$. The initial condition is $[x_1,x_2] = [0.5,0]$. We simulate the system on the time interval $[0,20]$ with sampling density $h = 0.05$. 

\subsubsection{Van der Pol Oscillator}

The dynamic of the Van der Pol oscillator follows the equation below 

\begin{equation}
\begin{aligned}
    \dot{x}_1 &= x_2\\
    \dot{x}_2 &= \mu (1 - x_1^2) x_2 -  x_1
\end{aligned},
\end{equation}
with $\mu = 0.5$. The initial condition is $[x_1,x_2] = [-2, 2]$. We simulate the system on the time interval $[0,10]$ with sampling density $h = 0.05$. 

\subsubsection{SEIR System}

The SEIR system is controlled by the equation 

\begin{equation}
\begin{aligned}
    \dot{S} &= -\beta S I\\
    \dot{E} &= \beta S I - \sigma E\\
    \dot{I} &=  \sigma E - \gamma I \\
    \dot{R} &= \gamma I 
\end{aligned},
\end{equation}
with parameter values chosen as $\beta = 0.3, \sigma = 0.2, \gamma = 0.1$. The initial condition is $[S,E,I,R] = [0.999,0.001,0,0]$. We simulate the system on the time interval $[0,160]$ with sampling density $h = 1.0$. 

\subsubsection{Lorenz 63 System}

The Lorenz 63 system is specified as follows:

\begin{equation}
\begin{aligned}
    \dot{x}_1 &=  \sigma(x_2 - x_1)\\
    \dot{x}_2 &= x_1 (\rho - x_3) - x_2\\
    \dot{x}_3 &= xy - \beta z \\
\end{aligned},
\end{equation}
with initial condition $x_1(0) = -8, \ x_2(0) = 7,\ x_3(0) = 27$. The coefficients are determined as $\rho = 28, \sigma = 10, \beta = 8/3$. We simulate the system on the time interval $[0,10]$ with sampling density $h = 0.05$. 

\subsubsection{R\"{o}ssler System}

The R\"{o}ssler system is governed by the nonlinear equation below

\begin{equation}
\begin{aligned}
    \dot{x}_1 &= -x_2 - x_3\\
    \dot{x}_2 &= x_1 + ax_2\\
    \dot{x}_3 &= b + x_3(x_1 - c) \\
\end{aligned},
\end{equation}
with initial condition $x_1(0) = -7.5, \ x_2(0) = 2.5, \ x_3(0) = 0$. The values of coefficients are set as $a = 0.2,b=0.2,c = 5.7$. We simulate the system on the time interval $[0,20]$ with sampling density $h = 0.05$.

\end{document}